\PassOptionsToPackage{unicode}{hyperref}
\PassOptionsToPackage{hyphens}{url}
\documentclass[
]{article}
\usepackage{xcolor}
\usepackage{amsmath,amssymb}
\usepackage{rotating}
\usepackage{iftex}
\ifPDFTeX
  \usepackage[T1]{fontenc}
  \usepackage[utf8]{inputenc}
  \usepackage{textcomp} % provide euro and other symbols
\else % if luatex or xetex
  \usepackage{unicode-math} % this also loads fontspec
  \defaultfontfeatures{Scale=MatchLowercase}
  \defaultfontfeatures[\rmfamily]{Ligatures=TeX,Scale=1}
\fi
\usepackage{lmodern}
\ifPDFTeX\else
\fi
\IfFileExists{upquote.sty}{\usepackage{upquote}}{}
\IfFileExists{microtype.sty}{% use microtype if available
  \usepackage[]{microtype}
  \UseMicrotypeSet[protrusion]{basicmath} % disable protrusion for tt fonts
}{}
\makeatletter
\@ifundefined{KOMAClassName}{% if non-KOMA class
  \IfFileExists{parskip.sty}{%
    \usepackage{parskip}
  }{% else
    \setlength{\parindent}{0pt}
    \setlength{\parskip}{6pt plus 2pt minus 1pt}}
}{% if KOMA class
  \KOMAoptions{parskip=half}}
\makeatother
\usepackage{longtable,booktabs,array}
\usepackage{calc} % for calculating minipage widths
\usepackage{etoolbox}
\makeatletter
\patchcmd\longtable{\par}{\if@noskipsec\mbox{}\fi\par}{}{}
\makeatother
\IfFileExists{footnotehyper.sty}{\usepackage{footnotehyper}}{\usepackage{footnote}}
\makesavenoteenv{longtable}
\usepackage{graphicx}
\makeatletter
\newsavebox\pandoc@box
\newcommand*\pandocbounded[1]{% scales image to fit in text height/width
  \sbox\pandoc@box{#1}%
  \Gscale@div\@tempa{\textheight}{\dimexpr\ht\pandoc@box+\dp\pandoc@box\relax}%
  \Gscale@div\@tempb{\linewidth}{\wd\pandoc@box}%
  \ifdim\@tempb\p@<\@tempa\p@\let\@tempa\@tempb\fi% select the smaller of both
  \ifdim\@tempa\p@<\p@\scalebox{\@tempa}{\usebox\pandoc@box}%
  \else\usebox{\pandoc@box}%
  \fi%
}
\def\fps@figure{htbp}
\makeatother
\NewDocumentCommand\citeproctext{}{}

\makeatletter
 \let\@cite@ofmt\@firstofone
 \def\@biblabel#1{}
 \def\@cite#1#2{{#1\if@tempswa , #2\fi}}
\makeatother
\newlength{\cslhangindent}
\newlength{\csllabelwidth}
\newenvironment{CSLReferences}[2] % #1 hanging-indent, #2 entry-spacing
 {\begin{list}{}{%
  \setlength{\itemindent}{0pt}
  \setlength{\leftmargin}{0pt}
  \setlength{\parsep}{0pt}
  \ifodd #1
   \setlength{\leftmargin}{\cslhangindent}
   \setlength{\itemindent}{-1\cslhangindent}
  \fi
  \setlength{\itemsep}{#2\baselineskip}}}
 {\end{list}}
\usepackage{calc}

\providecommand{\tightlist}{%
  \setlength{\itemsep}{0pt}\setlength{\parskip}{0pt}}
\usepackage{listings}
\lstdefinestyle{pythoncode}{
  language=Python,
  basicstyle=\ttfamily\footnotesize,
  keywordstyle=\color{blue!70!black},
  stringstyle=\color{orange!70!black},
  commentstyle=\color{green!50!black}\itshape,
  showstringspaces=false,
  breaklines=true,
  breakatwhitespace=true,
  frame=single,
  rulecolor=\color{gray!40},
  backgroundcolor=\color{gray!8},
  tabsize=4,
  xleftmargin=5pt,
  xrightmargin=5pt,
  aboveskip=6pt,
  belowskip=6pt,
}
\lstdefinestyle{output}{
  basicstyle=\ttfamily\footnotesize,
  showstringspaces=false,
  breaklines=true,
  frame=single,
  rulecolor=\color{gray!60},
  backgroundcolor=\color{gray!15},
  xleftmargin=5pt,
  xrightmargin=5pt,
  aboveskip=6pt,
  belowskip=6pt,
}
\usepackage{bookmark}
\IfFileExists{xurl.sty}{\usepackage{xurl}}{} % add URL line breaks if available
\hypersetup{
  pdftitle={OpenReservoirComputing: GPU-Accelerated Reservoir Computing in JAX},
  hidelinks,
  pdfcreator={LaTeX via pandoc}}

\usepackage{authblk}

\title{OpenReservoirComputing: GPU-Accelerated Reservoir Computing in JAX}
\author[1,*]{Jan P. Williams}
\author[1]{Dima Tretiak}
\author[1]{Steven L. Brunton}
\author[2,3,4]{J. Nathan Kutz}
\author[1]{Krithika Manohar}
\affil[1]{Department of Mechanical Engineering, University of Washington, USA}
\affil[2]{Department of Applied Mathematics, University of Washington, USA}
\affil[3]{Department of Electrical and Computer Engineering, University of Washington, USA}
\affil[4]{Autodesk Research, London, UK}
\affil[*]{Corresponding author}
\date{25 February 2026}

\begin{document}
\maketitle

\section{Summary}\label{summary}

OpenReservoirComputing (ORC) is a Python library for reservoir computing
(RC) written in JAX (Bradbury et al. 2018) and Equinox (Kidger and Garcia 2021). JAX is a Python
library for high-performance numerical computing that enables automatic
differentiation, just-in-time (JIT) compilation, and GPU/TPU
acceleration, while equinox is a neural network
framework for JAX. RC is a form of machine learning that
functions by lifting a low-dimensional sequence or signal into a
high-dimensional dynamical system and training a simple, linear readout
layer from the high-dimensional dynamics back to a lower-dimensional
quantity of interest. The most common application of RC is time-series
forecasting, where the goal is to predict a signal's future evolution.
RC has achieved state-of-the-art performance on this task, particularly
when applied to chaotic dynamical systems. In addition, RC approaches
can be adapted to perform classification and control tasks. ORC provides
both modular components for building custom RC models and built-in
models for forecasting, classification, and control. By building on JAX
and Equinox, ORC offers GPU acceleration, JIT compilation, and automatic
vectorization. These capabilities make prototyping new models faster and
enable larger and more powerful reservoir architectures. End-to-end
differentiability also enables seamless integration with other deep
learning models built with Equinox.

\section{Statement of Need}\label{statement-of-need}

Time-series prediction, classification, and control are fundamental
tasks across science and engineering, arising in applications from
climate modeling and fluid dynamics to robotics and neuroscience. Deep
learning approaches to these tasks typically require large datasets,
long training times, and expensive tuning of optimization
hyperparameters. RC offers a compelling alternative. Since only the
readout layer is trained via a single ridge regression, RC models can be
trained in a fraction of the time required by comparable recurrent
neural networks, often with less data and fewer hyperparameters to tune
(Lukoševičius and Jaeger 2009). This makes RC particularly attractive
for rapid prototyping, real-time applications, and settings where
training data is limited. However, realizing these benefits in practice
requires software that is both efficient and adaptable.

ORC's built-in models provide an easy entry point for users new to the
field. In particular, a new user can supply their own time-series data
to instantiate, train, and forecast in three simple lines of code.
Built-in visualization tools make it easy to evaluate model performance.
Varying the hyperparameters of built-in models lets users explore how RC
performance depends on configuration choices. While other RC libraries
allow for easy use with low-dimensional systems, ORC's JAX foundation
makes extending to higher-dimensional systems via parallel reservoirs
equally simple (also achievable in three lines of code). For
spatiotemporal forecasting problems, JAX's \texttt{vmap} transformation
enables efficient vectorization across parallel reservoirs that
decompose a spatial domain into overlapping subdomains, while JIT
compilation eliminates Python overhead in the reservoir state evolution
loop. Since RC approaches are so fast to train, this provides an easy
way for users to train lightweight surrogate models.

Much RC research is aimed at designing performant reservoir
architectures. ORC makes this easy through its use of abstract base
classes for Embedding, Driver, and Readout layers. Users need only
define forward pass logic to integrate a new reservoir topology or
readout strategy, while reusing the rest of the framework. This modular
approach also allows for easy ablation studies comparing how different
components affect RC performance. It also makes it straightforward to
reuse existing architectures, without the additional complexity of
writing teacher forcing or autoregressive prediction functions from
scratch.

Because of ORC's functional approach in JAX, built-in and user-created
models provide end-to-end differentiability by default. This enables
gradient-based optimization of input sequences for control problems.
This also makes ORC well suited to integrate with deep learning models
such as those presented in (Özalp et al. 2023, 2025). ORC models are
simply Equinox modules, allowing them to be composed with other Equinox
models such as standard neural networks, neural operators (Kidger and
Garcia 2021), etc.

\section{State of the Field}\label{state-of-the-field}

\begin{table}[htbp]
\centering

\label{tbl:comparison}
\resizebox{\textwidth}{!}{%
\begin{tabular}{lllccccccc}
\toprule
 & Lang. & GPU & Auto.\ Diff. & Parallel./Vect. & Forecast. & Classif. & Control & Cont.\ Time \\
\midrule
\textbf{ORC}                   & Python & \checkmark  & \checkmark & \checkmark/\checkmark & \checkmark & \checkmark & \checkmark & \checkmark \\
\textbf{ReservoirPy}           & Python & \checkmark* & $\times$   & $\times$/$\times$     & \checkmark & \checkmark & $\times$   & $\times$   \\
\textbf{ReservoirComputing.jl} & Julia  & \checkmark  & \checkmark & \checkmark/$\times$   & \checkmark & \checkmark & $\times$   & $\times$   \\
\bottomrule
\end{tabular}%
}
\caption{Comparison of reservoir computing libraries across key features.
\checkmark~indicates full support; $\times$ indicates no support.
*ReservoirPy's GPU support is available via its JAX backend (v0.4.0+)
but does not fully exploit JAX's functional programming model.
\emph{Parallelizable} denotes native support for parallel RC
architectures as in (Pathak et al. 2018) and \emph{vectorizable} denotes
native support for vectorization (e.g.\ \texttt{vmap}).}
\end{table}

The most commonly used open-source library for reservoir computing is
ReservoirPy (Trouvain et al. 2020). Much like ORC, ReservoirPy provides
a variety of built-in architectures, as well as an easy-to-use API for
designing one's own layers. ReservoirPy was initially built on NumPy and
SciPy with the maintainers adding a JAX backend in v0.4.0. However, ORC
differs from ReservoirPy in several important ways.

First, ORC was \emph{designed} on top of JAX (Bradbury et al. 2018) and
Equinox (Kidger and Garcia 2021), which provide a functional programming
model for numerical computing. JAX's composable transformations include
\texttt{vmap} (automatic vectorization, which maps a function over
batched inputs without explicit loop code), \texttt{grad} (automatic
differentiation for computing gradients), and \texttt{jit} (just-in-time
compilation to hardware-accelerated code). These cannot be retrofitted
into a NumPy-based architecture; NumPy's imperative, stateful design
relies on mutable objects and side effects that are incompatible with
JAX's requirement for pure functions. While the JAX backend of
ReservoirPy does improve performance, the API cannot fully exploit JAX's
capabilities for this reason. As an example, ORC's autoregressive forecast loop uses
\texttt{jax.lax.scan} (a JIT-compatible loop primitive with explicit
carry state that avoids the Python-level overhead of a
\texttt{for} loop), whereas ReservoirPy's object-oriented design makes
this impossible. Moreover, the compatibility of ORC models with
\texttt{vmap} makes the implementation of ensembles efficient and easy.

Second, ORC has a different built-in feature set. ORC supports
continuous-time reservoir dynamics via Diffrax (Kidger 2021), a
JAX-based library for solving differential equations with adaptive-step
integrators, allowing users to define reservoir equations as ordinary
differential equations. ORC also supports novel
architectures such as Taylor-expanded and GRU-based drivers alongside
standard echo state networks. ReservoirPy does not provide these
functionalities.

Third, ORC models work seamlessly with other deep learning models
implemented in Equinox. ReservoirPy is an outstanding library for
standalone RC tasks that do not need to integrate with other deep
learning frameworks. Working with NumPy rather than JAX may also be more
accessible for users unfamiliar with functional programming. However,
ORC's design priorities and advantages are nonetheless distinct from
ReservoirPy's, as outlined above. ORC's syntax also is more consistent
with modern deep learning libraries.

Other open source libraries for RC include Pytorch-ESN (Nardo 2018) and
ReservoirComputing.jl (Martinuzzi et al. 2022). Pytorch-ESN allows for
integration with other PyTorch models, but is not as widely adopted and
offers a more limited feature set than ORC or ReservoirPy.
ReservoirComputing.jl is a widely used Julia library for RC that
influenced many of ORC's design choices. In particular, ORC's modular
design draws heavily from ReservoirComputing.jl. Since Python dominates
much of machine learning research, bringing these capabilities to Python
is valuable for the broader ML community. A summary of the functionality
of ORC, ReservoirPy, and ReservoirComputing.jl is presented in Table 1.

\section{Software Design}\label{software-design}

\begin{figure}
\centering
\pandocbounded{\includegraphics[keepaspectratio]{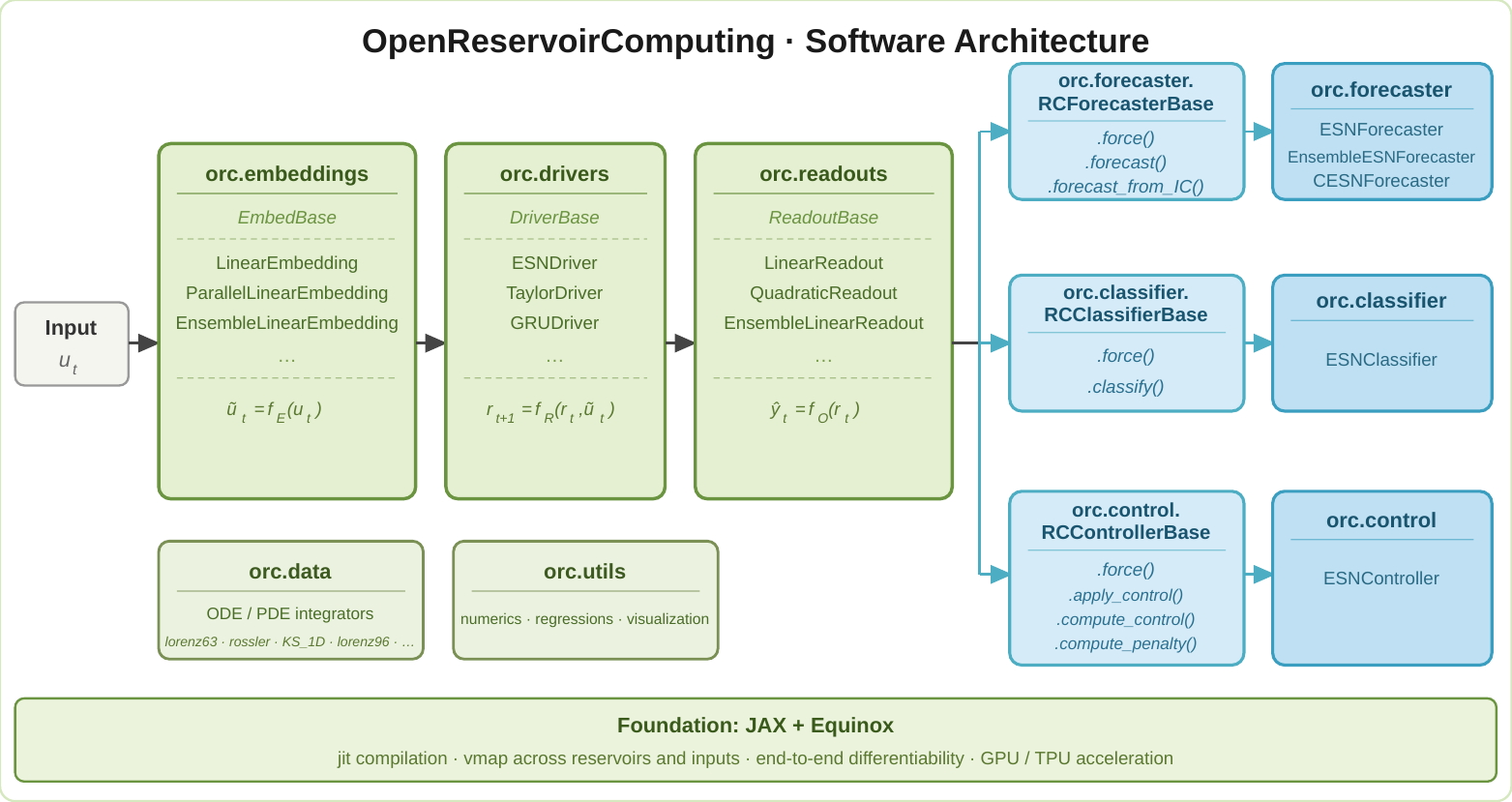}}
\caption{ORC three-layer pipeline architecture. Each reservoir computer (RC) is decomposed into (i) an ``embedding'' function that lifts a low-dimensional signal $u_t$ to the reservoir dimension, (ii) a ``driver'' function that propagates the reservoir state, and (iii) a ``readout'' that maps back to a target $y$. For control and forecasting RCs, the target $y$ is typically $u_{t+1}$ (either in the presence of a forcing term or not) and for classification the target $y$ is a label. 
\label{fig:architecture}}
\end{figure}

ORC models are decomposed into three components, illustrated in
\autoref{fig:architecture}: (i) an embedding \(f_E\) that lifts a
low-dimensional input signal \(u_t\) to a high-dimensional space, (ii) a
driver \(f_R\) that propagates the high-dimensional state \(r_t\), and
(iii) a readout \(f_O\) that maps the latent state back to an
approximation of some low-dimensional signal \(y_t\). Depending on the
task at hand, \(y_t\) may be a future time-step of \(u_t\), a label
associated with input data, or some other target signal. ORC differs
from many existing approaches that unify (i) and (ii). Separating the
embedding from the reservoir state propagation allows for cleaner
application of RC to non-standard tasks, such as acting as a surrogate
model for model predictive control. It also allows for the easier
incorporation of non-standard embeddings, including stochastic
embeddings that can arise in the study of physical RC systems. Moreover,
the modular design lets components developed for one task (e.g.,
forecasting) be reused directly in another (e.g., classification).

All components are implemented as Equinox modules (Kidger and Garcia
2021), which are immutable pytree-registered objects. Model parameters
(reservoir weights, readout matrices) are stored as JAX arrays within
the module, and parameter updates produce new module instances via
\texttt{eqx.tree\_at} rather than mutating the state in place. This
functional design enables JAX's composable transformations to operate
directly on model objects and allows ORC models to be composed with
other Equinox modules.

ORC supports parallel reservoirs (Pathak et al. 2018) by default via a
\texttt{chunks} parameter in each module. This slightly complicates
built-in training support and introduces an extra tensor dimension that
can make the API less intuitive at first. However, we believe this
tradeoff is worthwhile because it enables spatiotemporal RC methods
unavailable in other libraries. The inclusion of this tensor dimension
also allows for simpele batching of parallel reservoirs during training,
avoiding excessive GPU VRAM requirements.

ORC provides unified training functions that work
with any model inheriting from the corresponding base class, including
user-defined models with custom components. These functions delegate
shape handling to the readout layer and accept keyword arguments that
are forwarded to the model's \texttt{force} method, allowing the same
training function to handle both discrete and continuous-time models
transparently.

The library provides three built-in model classes:
\texttt{ESNForecaster} for time series prediction,
\texttt{ESNClassifier} for sequence classification, and
\texttt{ESNController} for learning control policies with exogenous
control inputs. Each composes embedding, driver, and readout components
and provides task-specific methods (\texttt{forecast},
\texttt{classify}, \texttt{apply\_control}). Users who need custom
architectures can subclass the abstract base classes, define only the
components that differ, and immediately use the unified training
functions without reimplementing teacher forcing, autoregressive
prediction, or ridge regression.

ORC also includes a data generation module with ODE and PDE integrators
for standard benchmark systems, including the Lorenz-63 attractor,
Rössler system, double pendulum, Lorenz-96 model, and the
Kuramoto-Sivashinsky equation, all implemented using Diffrax.

\section{Research Impact Statement}\label{research-impact-statement}

\begin{figure}
\centering
\pandocbounded{\includegraphics[keepaspectratio,alt={ORC with GPU acceleration enables significantly faster performance than ReservoirPy, even when using ReservoirPy's JAX backend. Panel (a) shows the time per forecast step of RC models with varying reservoir dimension trained to forecast the Lorenz system, while panel (b) shows training tine for RC models with fixed reservoir dimension of 2000 but varying number of training samples. Performance of the two libraries with and without GPU acceleration are shown. GPU results were obtained running on an NVIDIA A40 GPU and CPU results were obtained with an Apple M2 chip. }]{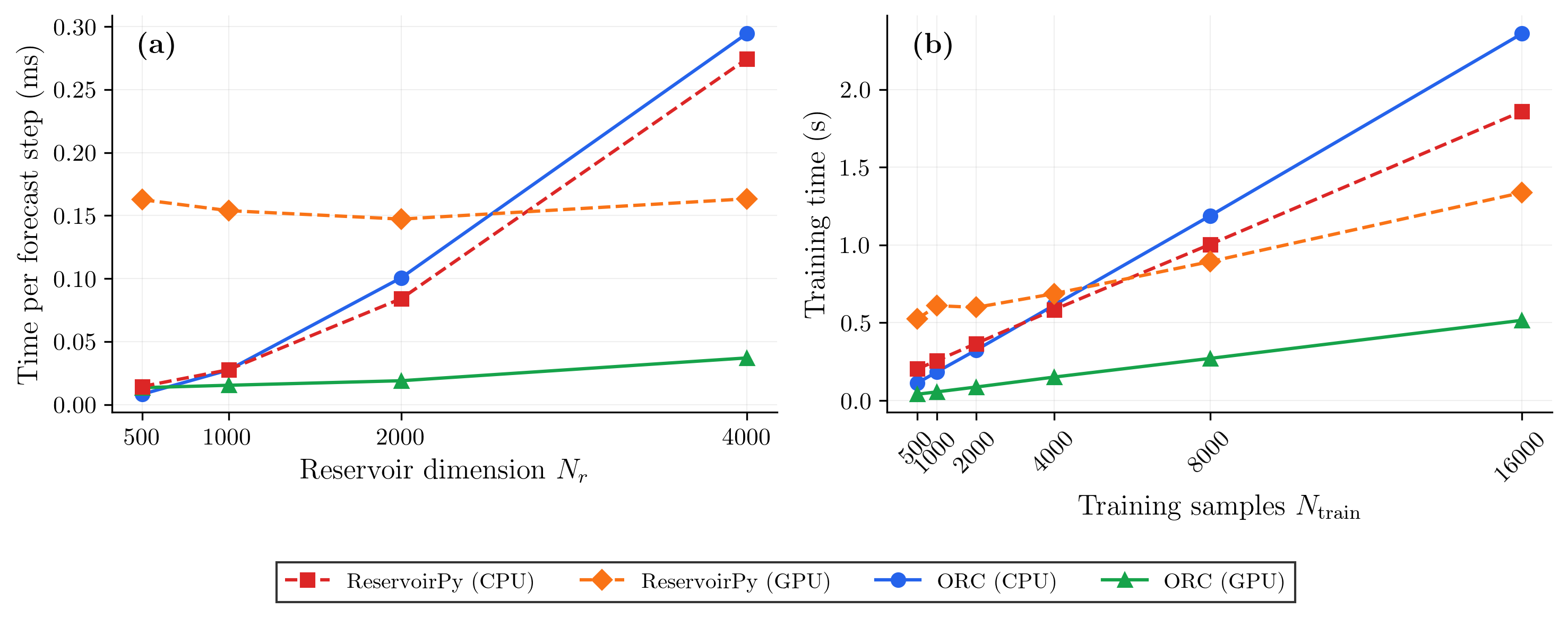}}
\caption{ORC with GPU acceleration enables significantly faster
performance than ReservoirPy, even when using ReservoirPy's JAX backend.
Panel (a) shows the time per forecast step of RC models with varying
reservoir dimension trained to forecast the Lorenz system, while panel
(b) shows training tine for RC models with fixed reservoir dimension of
2000 but varying number of training samples. Performance of the two
libraries with and without GPU acceleration are shown. GPU results were
obtained running on an NVIDIA A40 GPU and CPU results were obtained with
an Apple M2 chip. \label{fig:comp}}
\end{figure}

ORC addresses an immediate need for RC researchers; it allows for easy
reimplementation of architectures such as (Özalp et al. 2023, 2025) that
integrate reservoir computing with larger neural network architectures,
and makes it easier to iterate on these ideas than any existing library.
We also benchmark ORC against ReservoirPy in
\autoref{fig:comp}, across reservoir sizes and number of training samples. We find that with GPU
acceleration, ORC scales far more favorably than ReservoirPy. Moreover,
ORC is the only current package that supports training RC for control
tasks, thanks to its efficiency and the end-to-end differentiability
that JAX and Equinox provide. RC-based control has recently been shown
to be advantageous over other popular forms of RNN-based MPC (Williams
et al. 2024). ORC's performance also makes it well suited for the
parallel architectures needed to extend RC to higher-dimensional
data.

\section{AI Usage Disclosure}\label{ai-usage-disclosure}

Claude (Anthropic) was used for code assistance during code debugging,
proofreading this paper, and the generation of the architecture
visualization SVG. ChatGPT (OpenAI) was used to generate the ORC logo.
All generated code was reviewed, tested, and validated by the authors.

\section{Acknowledgements}\label{acknowledgements}

The authors acknowledge support from the National Science Foundation AI
Institute in Dynamic Systems (grant number 2112085). The authors also
thank Anastasia Bizyaeva, Noa Kaplan, Ling-Wei Kong for insightful
conversations.

\protect\phantomsection\label{refs}

\appendix
\pagebreak
\section{Appendix A: Reservoir Computing Background}\label{appendix:rc-background}
The purpose of this notebook is to provide a broad introduction to
reservoir computing (RC) by walking through the creation of an RC model
within the ORC framework. RC was initially proposed independently by
Maas, Natschläger, and Markram {[}1{]}, and Jaeger {[}2{]} in the early
2000s, largely as a means of avoiding issues with training recurrent
neural networks (RNNs) via backpropagation and gradient descent. Since
then, the field of RC has exploded, introducing many RC inspired
architectures and new applications of RC approaches.

One of the central premises of ORC is that the vast majority of RC
architectures can be decomposed into three components: an embedding that
lifts a low-dimensional input signal \(u_t\) to a high-dimensional
space, a driver that propagates a high-dimensional latent state \(r_t\),
and a readout that maps the latent state back to a lower dimensional
signal \(\hat y_t\). More concretely, we can typically represent the
embedding \(f_E\), driver \(f_R\), and readout \(f_O\) as functions1.
Schematically, a reservoir can be illustrated with these functions as a
part of the below control loop. The shaded blue boxes are instantiated
independent of training data, greatly simplifying the training of
\(f_O\).

\pandocbounded{\includegraphics[keepaspectratio,alt={image.png}]{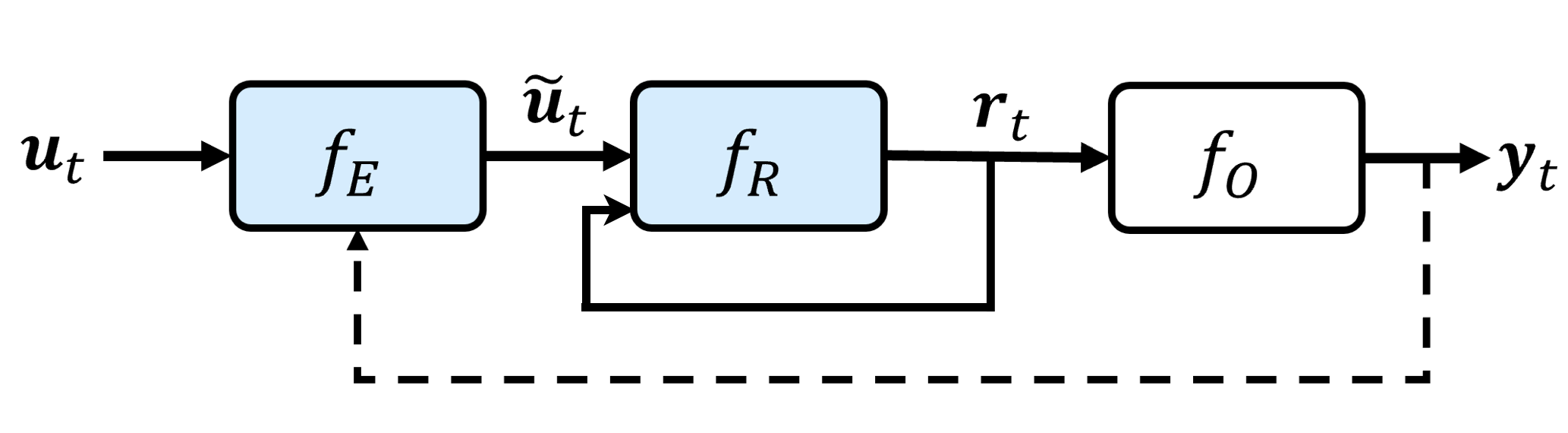}}

More formally, each function can be defined:

\[f_{E} : \mathbb R^{N_u} \to \mathbb R^{N_r}  \quad \text{with} \quad \tilde u_{t} = f_{E}(u_t),\quad \text{(embedding)}\]
\[f_{R} : \mathbb R^{N_r} \times \mathbb R^{N_r} \to \mathbb R^{N_r} \quad \text{with} \quad r_{t+1} = f_R(r_t, \tilde u_t), \quad \text{(driver)}\]
and
\[f_{O} : \mathbb R^{N_y} \to \mathbb R^{N_r} \quad \text{with} \quad \hat y_{t} = f_O(r_t). \quad \text{(readout)}\]

We\textquotesingle ll call \(N_u\) the input dimension, \(N_r\) the
reservoir dimension, and \(N_y\) the output dimension. RC makes the
assumption that if the reservoir dimension is sufficiently large
(\(N_r \gg N_u\)) and the dynamics of the driver are sufficiently
"expressive," then the parameters/weights of both the embedding and
driver can be set independent of any training data; only the readout
layer is trained. This greatly simplifies the training of an RC in
comparison to RNNs as it completely avoids backpropagation through time.
It also allows for the implementation of a wider variety of embeddings
and drivers, including physical implementations of RC.

Throughout the rest of this notebook, we illustrate
how ORC can be used to create and train a RC for forecasting with
user-defined embedding, driver, and readout functions. For this example,
we use a discrete time reservoir, however, ORC is also capabable of
using continuous time drivers as will be discussed in a future tutorial.

\begin{lstlisting}[style=pythoncode]
# imports
import equinox as eqx
import jax
import jax.numpy as jnp

import orc
from orc.utils.regressions import ridge_regression
\end{lstlisting}

\subsection{Defining an embedding}\label{defining-an-embedding}

By far, the most common choice for embedding function is to assume the
form \(f_{E}(u) = W_{E} u\) where
\(W_{E} \in \mathbb R ^{N_r \times N_u}\) with entries drawn
independently according to some distribution. However, to illustrate
that ORC allows for a very broad range of embeddings,
we\textquotesingle ll take \(f_{E}\) to be a two-layer MLP with an ELU
activation function. To do so, we\textquotesingle ll inherit from
\texttt{orc.embeddings.EmbedBase}, which expects us to specify an input
dimension \texttt{in\_dim}, a reservoir dimension \texttt{res\_dim}, and
define a method \texttt{embed} that maps from \texttt{in\_dim} to
\texttt{res\_dim}.

\begin{lstlisting}[style=pythoncode]
# define embedding function
class ELUEmbedding(orc.embeddings.EmbedBase):

    W1: jnp.ndarray  # weight matrix for first layer of ELU MLP
    W2: jnp.ndarray  # weight matrix for second layer of ELU MLP
    b1: jnp.ndarray  # bias for first layer of ELU MLP
    b2: jnp.ndarray  # bias for second layer of ELU MLP

    def __init__(self, in_dim, res_dim, seed=0):
        super().__init__(in_dim=in_dim, res_dim=res_dim)
        rkey = jax.random.key(seed)
        W1key, W2key, b1key, b2key = jax.random.split(rkey, 4)
        # random initialization of parameters of ELU MLP
        self.W1 = jax.random.normal(
            W1key, shape=(res_dim // 2, in_dim)
        )
        self.W1 = self.W1 / jnp.sqrt((res_dim // 2) * in_dim)
        self.W2 = jax.random.normal(
            W2key,
            shape=(res_dim, res_dim // 2)
        )
        self.W2 = self.W2 / jnp.sqrt(res_dim * (res_dim // 2))
        self.b1 = jax.random.normal(
            b1key,
            shape=(res_dim // 2)
        )
        self.b1 = self.b1 / jnp.sqrt(res_dim // 2)
        self.b2 = jax.random.normal(b1key, shape=(res_dim))
        self.b2 = self.b2 / jnp.sqrt(res_dim)

    def embed(self, in_state):
        in_state = self.W1 @ in_state + self.b1
        in_state = jax.nn.elu(in_state)
        in_state = self.W2 @ in_state + self.b2
        in_state = jax.nn.elu(in_state)
        return in_state
\end{lstlisting}

\subsection{Defining a driver}\label{defining-a-driver}

For the driver, we use the update equations of a
gated recurrent unit (GRU). The GRU is already implemented in
\texttt{equinox} so we need only wrap the existing \texttt{eqx.Module}
in a \texttt{DriverBase} class. Although we redefine it here for
illustrative purposes, this driver is actually built into \texttt{ORC}.

\begin{lstlisting}[style=pythoncode]
# define driver function
class GRUDriver(orc.drivers.DriverBase):

    gru: eqx.Module

    def __init__(self, res_dim, seed=0):
        super().__init__(res_dim=res_dim)
        key = jax.random.key(seed)
        self.gru = eqx.nn.GRUCell(res_dim, res_dim, key=key)

    def advance(self, res_state, in_state):
        return self.gru(in_state, res_state)
\end{lstlisting}

\subsection{Defining a readout}\label{defining-a-readout}

While there is great flexibility in the assumed form of \(f_E\) and
\(f_R\), \(f_O\) is more restricted if we want to retain the ability to
train the RC with a linear regression. We\textquotesingle ll assume the
output operator is linear, that is \(f_O(r_t) = W_Or_t\) where
\(W_O \in \mathbb R^{N_y \times N_r}\). We initialize the readout layer
of our RC by subclassing \texttt{orc.readouts.ReadoutBase} and defining
a \texttt{readout} method. We\textquotesingle ll initialize \(W_O\) with
zeros, although the initialization won\textquotesingle t affect
training.

\begin{lstlisting}[style=pythoncode]
# define readout function
class Readout(orc.readouts.ReadoutBase):
    res_dim: int
    out_dim: int
    wout: jnp.ndarray

    def __init__(self, out_dim, res_dim):
        super().__init__(out_dim, res_dim)
        self.wout = jnp.zeros((out_dim, res_dim))

    def readout(self, res_state):
        return self.wout @ res_state
\end{lstlisting}

\subsection{Defining a forecaster}\label{defining-a-forecaster}

We\textquotesingle re now able to combine these three components into a
\texttt{Forecaster} that inherits from \texttt{orc.rc.RCForecasterBase}.

\begin{lstlisting}[style=pythoncode]
# define RC as embedding + driver + readout
class Forecaster(orc.forecaster.RCForecasterBase):
    driver: orc.drivers.DriverBase
    readout: orc.readouts.ReadoutBase
    embedding: orc.embeddings.EmbedBase
\end{lstlisting}

\subsection{Training}\label{training}

Now that the input, driver, and readout functions are defined and
pacakged into a \texttt{Forecaster} object, we can turn to the training
procedure for the model. For some example data, we\textquotesingle ll
use a numerical integration of the Rossler system.

\begin{lstlisting}[style=pythoncode]
# integrate Rossler system
tN = 200
dt = 0.01
u0 = jnp.array([-10, 2, 1], dtype=jnp.float64)
U,t = orc.data.rossler(tN=tN, dt=dt, u0=u0)
split_idx = int(U.shape[0] * 0.8)
U_train = U[:split_idx]
U_test = U[split_idx:]
t_test = t[split_idx:]
orc.utils.visualization.plot_time_series(
    U,
    t,
    state_var_names=["$u_1$", "$u_2$", "$u_3$"],
    title="Rossler Data",
    x_label= "$t$",
)
\end{lstlisting}

\pandocbounded{\includegraphics[keepaspectratio]{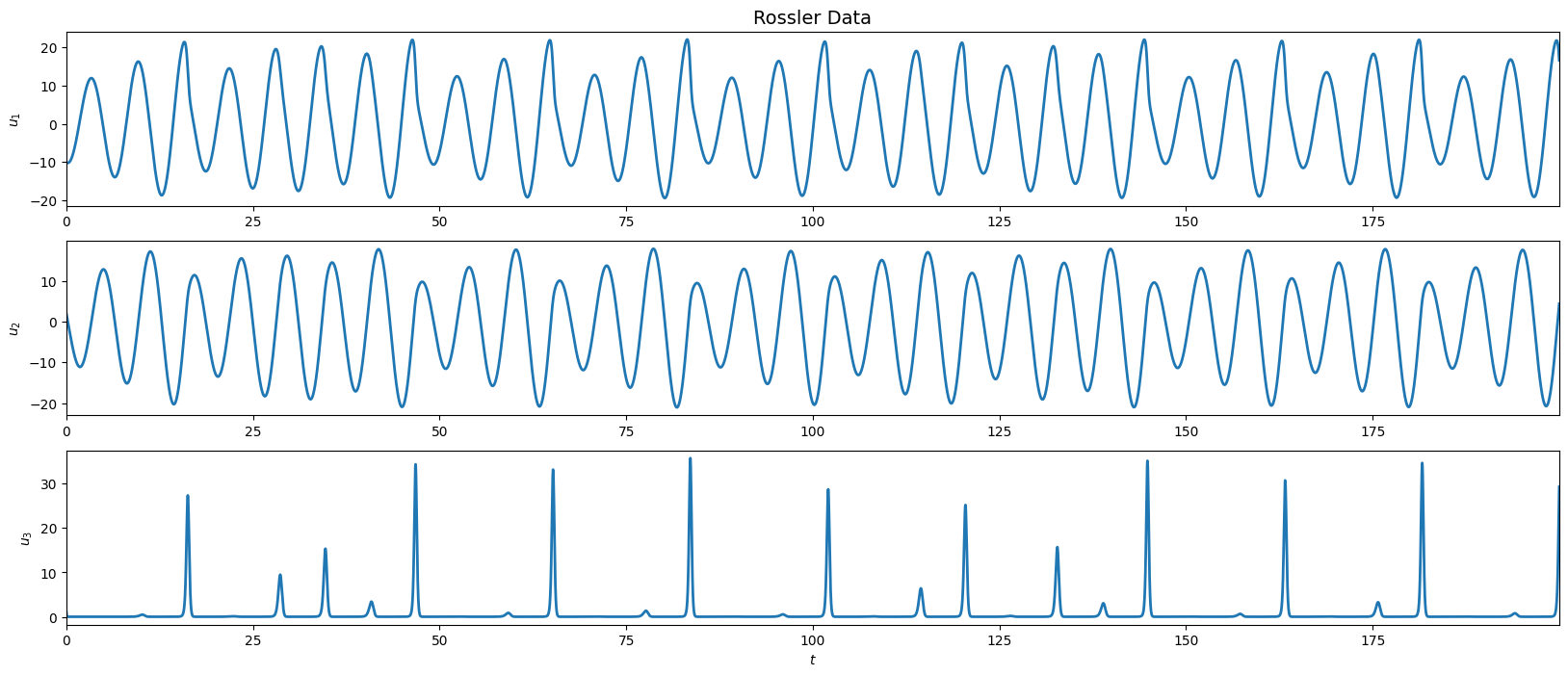}}

The desired input (and output) dimension is that of the Rossler system,
\(N_y = N_u = 3\). We\textquotesingle ll take \(N_R\) = 500 for this
example.

\begin{lstlisting}[style=pythoncode]
# create RC
Nr = 500
Nu = 3
driver = GRUDriver(Nr)
embedding = ELUEmbedding(Nu, Nr)
readout = Readout(Nu, Nr)
model = Forecaster(driver, readout, embedding)
\end{lstlisting}

Our training data can be represented in a data matrix as \[
U_{\text{data}} = \begin{bmatrix}
| & | & & |\\
u_1 & u_1 & \cdots & u_T \\
| & | &  & |
\end{bmatrix}.
\] Now, setting \(r_0 = \mathbf 0\), we can compute
\(r_1 = f_R(r_0, f_E(u_0))\), \(r_2 = f_R(r_1, f_E(u_1))\), \(\dots\),
\(r_T = f_R(r_{T-1}, f_E(u_{T-1}))\) to generate \[
R_{\text{data}}\begin{bmatrix}
| & | & & |\\
r_1 & r_1 & \cdots & r_T \\
| & | &  & |
\end{bmatrix}.
\] Learning \(W_O\) now just amounts to performing a Ridge regression to
satisfy \[R_{data}^T W_{O}^T = U_{data}^T.\] In practice, we discard the
first \texttt{spinup} columns of \(R_{data}\) and \(U_{data}\) to remove
the effect of our (arbitrary) choice \(r_0 = \mathbf 0.\) The ability to
discard this initial transient relies on the principle of generalized
synchronization and is frequently referred to as the echo state property
in RC literature.

Since we\textquotesingle ve inherited from \texttt{RCForecasterBase},
generating \(R_{data}\) is easily accomplished in a single line.

\begin{lstlisting}[style=pythoncode]
# teacher force the reservoir
forced_seq = model.force(U_train[:-1], res_state=jnp.zeros((Nr)))

# shift the indices of the target sequence of training data
target_seq = U_train[1:]

# set transient to discard
spinup = 200

# learn wout
readout_mat = ridge_regression(
    forced_seq[spinup:],
    target_seq[spinup:],
    beta=1e-7
)
\end{lstlisting}

We\textquotesingle d now like to update the \texttt{readout} attribute
of \texttt{model} to be equal to the learned \texttt{readout\_mat}.
However, under the hood many ORC objects are instances of
\texttt{equinox.Module}, which are immutable. Thus, we need to create a
new \texttt{Forecaster} object with \texttt{readout.W\_O} set to
\texttt{readout\_mat}. This can be done with
\texttt{equinox}\textquotesingle s convenient \texttt{tree\_at}
function.

\begin{lstlisting}[style=pythoncode]
# define where in the forecaster model we need to update
def where(model: Forecaster):
    return model.readout.wout
model = eqx.tree_at(where, model, readout_mat)
\end{lstlisting}

\begin{lstlisting}[style=pythoncode]
# Alternatively
model, forced_seq = orc.forecaster.train_RCForecaster(
    model,
    train_seq=U_train,
    spinup=spinup,
    beta=1e-7
)
\end{lstlisting}

We\textquotesingle re now ready to perform a forecast and evaluate our
trained RC:

\begin{lstlisting}[style=pythoncode]
# perform forecast
U_pred = model.forecast_from_IC(
    fcast_len=U_test.shape[0],
    spinup_data=U_train[-spinup:]
)

# plot forecast
orc.utils.visualization.plot_time_series(
    [U_test, U_pred],
    (t_test - t_test[0]) * 0.07,
    state_var_names=["$u_1$", "$u_2$", "$u_3$"],
    time_series_labels=["True", "Predicted"],
    line_formats=["-", "r--"],
    x_label= r"$\lambda _1 t$",
)
\end{lstlisting}

\pandocbounded{\includegraphics[keepaspectratio]{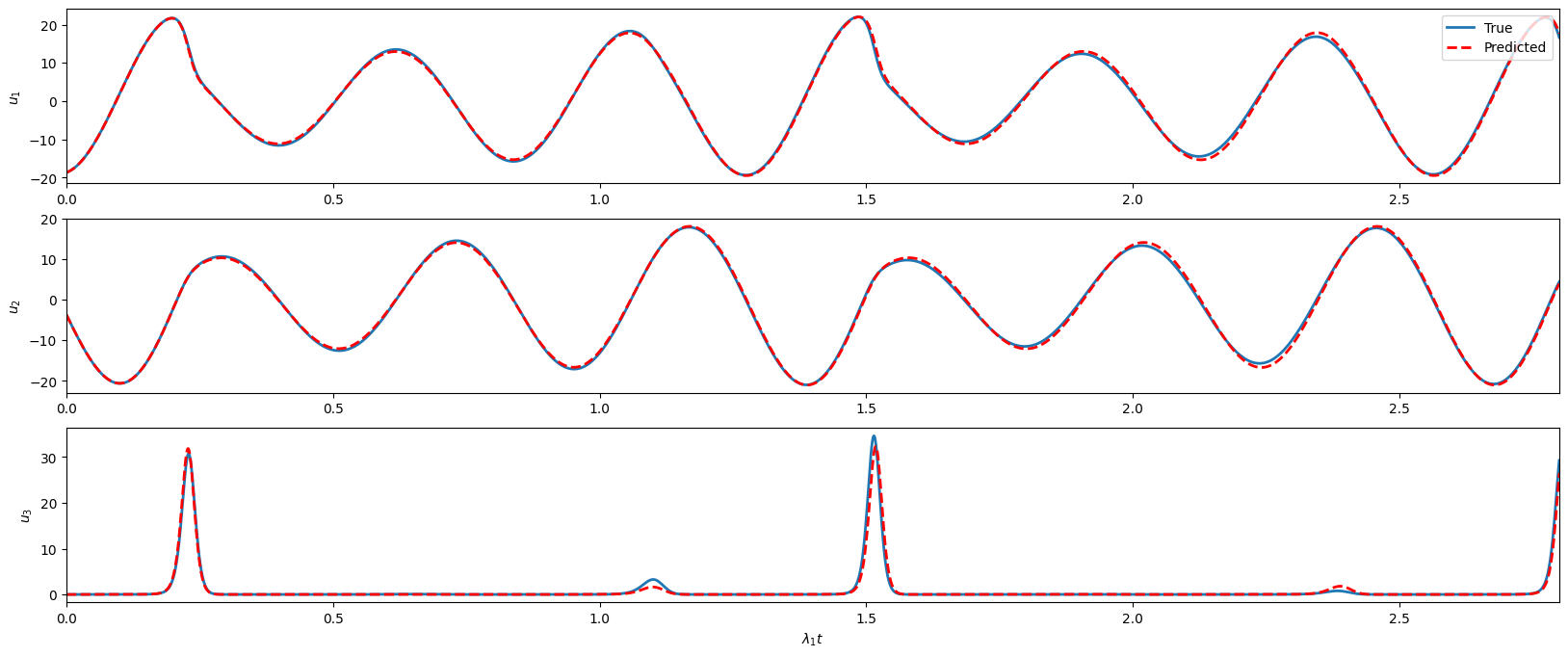}}

Let\textquotesingle s summarize what we\textquotesingle ve seen in this
notebook. We first defined an embedding, driver, and form of readout. We
then packaged these together in a \texttt{Forecaster} that allowed us to
easily train and evaluate the model architecture on the chaotic Rossler
system. Provided that the input/output and reservoir dimensions were
held fixed, we could easily switch out any of the components of the
\texttt{Forecaster} for different forms/mappings (either from ORC or by
defining new ones ourselves!). This modularity is a key benefit of ORC,
especially when taken with the ease of defining new mappings.

\subsection{References and footnotes}\label{references-and-footnotes}

{[}1{]} Maas, Natschläger, and Markram, "Real-time computing without
stable states: A new framework for neural computation based on
perturbations," \emph{Neural Computation}, 2002.

{[}2{]} Jaeger, "The ``echo state'' approach to analysing and training
recurrent neural networks-with an erratum note," \emph{German national
research center for information technology gmd technical report}, 2001.

1 There is growing interest in the field of stochastic reservoir
computers, especially quantum inspired RC, where one cannot always
define the embedding, driver, and readout as functions.

\section{Appendix B: Lorenz Forecasting Example}\label{appendix:lorenz}
This notebook illustrates some of the core built-in features offered by
ORC. Specifically, we train an echo state network (ESN) to forecast the
Lorenz63 system. An ESN is a recurrent neural network with a fixed,
random reservoir --- only the output (readout) layer is trained, making
it fast and simple to fit. Forecasting low-dimensional, chaotic dynamics
is a canonical test of reservoir computer (RC) performance and various
RC architectures have achieved state-of-the-art performance at this
task. The tutorial will also demonstrate the functionality of
ORC\textquotesingle s data module for integrating test systems for new
forecasting algorithms.

Only two imports are necessary to run the examples in this notebook:
\texttt{jax} and \texttt{orc} itself. Although not a strict requirement,
it is \textbf{strongly recommended} to use 64 bit floats when using ORC
and by default FP64 is enabled. Higher precision numerics are often
helpful in training accurate forecasting models.

\begin{lstlisting}[style=pythoncode]
import jax
import jax.numpy as jnp
import orc
\end{lstlisting}

\subsection{Generate Lorenz data}\label{generate-lorenz-data}

For user convenience, \texttt{orc.data} provides a variety of ODE and
PDE test systems, including the Lorenz system. The user must simply
specify the time to integrate, \texttt{tN}, and the discretization,
\texttt{dt}. Here, we also provide an initial condition,
\(u_0 = \begin{bmatrix} -10 & 1 & 10\end{bmatrix}^T\), although a
default initial condition can be used by omitting the argument
\texttt{u0}. Runnning the integration generates \texttt{U} the solution
array with shape \texttt{(int(tN\ /\ dt),\ 3)} and the corresponding
timesteps of the solution array, \texttt{t}.

\begin{lstlisting}[style=pythoncode]
# integrate
tN = 100
dt = 0.01
u0 = jnp.array([-10, 1, 10], dtype=jnp.float64)
U,t = orc.data.lorenz63(tN=tN, dt=dt, u0=u0)
\end{lstlisting}

\texttt{orc.utils.visualization} provides a convenient function for
plotting time-series, allowing us to visualize the integration we just
performed.

\begin{lstlisting}[style=pythoncode]
# plot numerical integration
orc.utils.visualization.plot_time_series(
    U,
    t,
    state_var_names=["$u_1$", "$u_2$", "$u_3$"],
    title="Lorenz63 Data",
    x_label= "$t$",
)
\end{lstlisting}

\pandocbounded{\includegraphics[keepaspectratio]{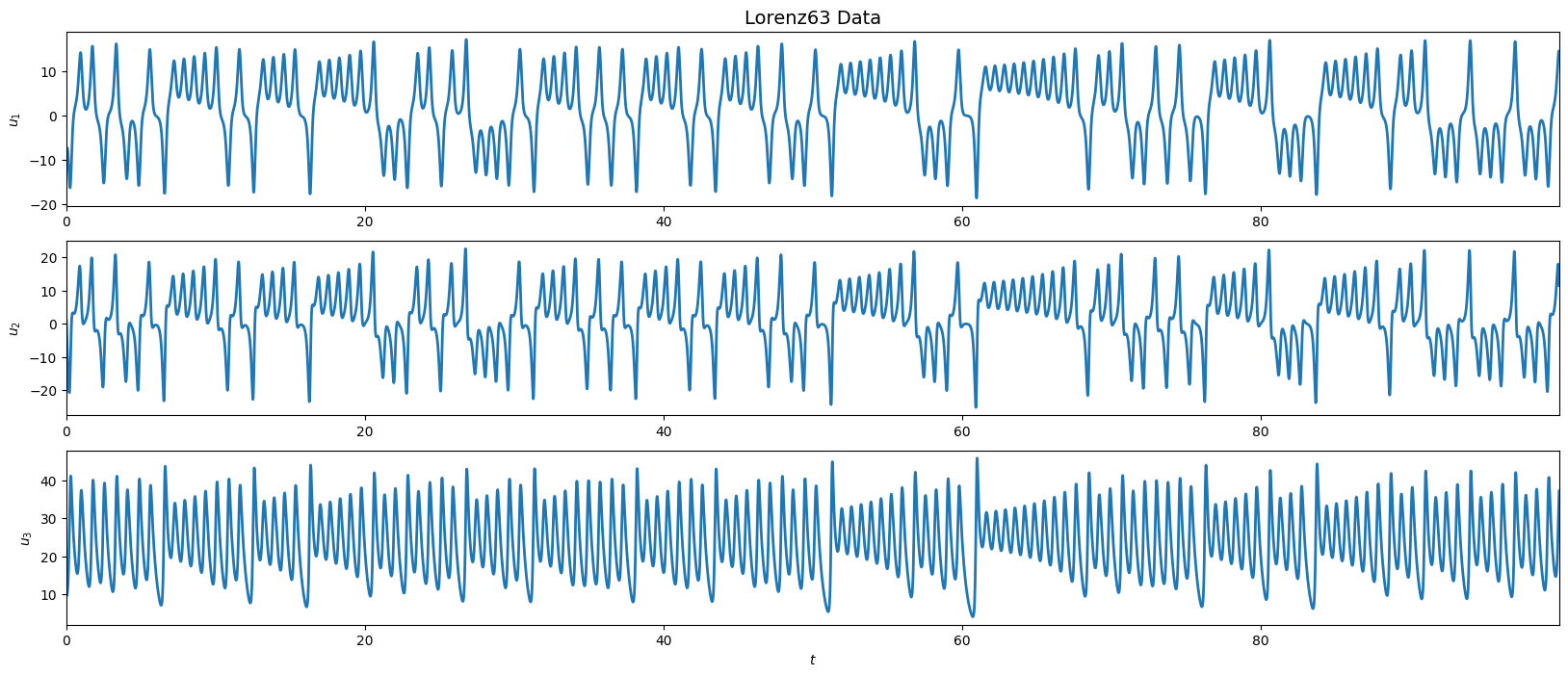}}

We\textquotesingle ll use the first 80\% of this trajectory to train our
ESN model and test its performance on the following 20\%.

\begin{lstlisting}[style=pythoncode]
# train-test split
test_perc = 0.2
split_idx = int((1 - test_perc) * U.shape[0])
U_train = U[:split_idx, :]
t_train = t[:split_idx]
U_test = U[split_idx:, :]
t_test = t[split_idx:]
test_timesteps = t_test.shape[0]
\end{lstlisting}

The simplest entry point for getting started with training RC models in
ORC is to use one of the built in models available in
\texttt{orc.forecaster}. Here, we use \texttt{ESNForecaster} which is an
implementation of an ESN following the conventions laid out in {[}1{]}.
\texttt{ESNForecaster} allows the user to set a variety of
hyperparameters, but the defaults work well for the Lorenz system. Key
hyperparameters include \texttt{res\_dim} (the number of reservoir
neurons --- larger reservoirs have more capacity but are slower to
train) and \texttt{Wr\_spectral\_radius} (controls the memory of the
reservoir --- values near 1.0 give longer memory). To instantiate the
model, the user need only specify the dimensionality of the data (in
this case, 3), the desired reservoir or latent state dimension, and an
integer random seed for the network.

Training is performed by the function
\texttt{orc.forecaster.train\_ESNForecaster}, which accepts as arguments
an \texttt{ESNForecaster} and time-series of training data
\texttt{U\_train} with shape \texttt{(N\_train,\ data\_dim)}. The
training procedure teacher-forces the reservoir --- driving it with the
ground truth input sequence to build up internal states --- then fits
the readout layer via ridge regression. The trained model as well as the
sequence of training reservoir states are returned. The latter of which
has the shape \texttt{(N\_train,\ chunks,\ res\_dim)} where in this case
\texttt{chunks\ =\ 1.}

The additional dimension in the sequence of reservoir states is a hint
that \texttt{ESNForecaster} models allow for the easy propagation of
\emph{parallel} reservoirs. Although not necessary in lower dimensional
cases like the Lorenz system, this parallelizability is helpful in
applying RC methods to higher dimensional systems. The example notebook
\textbf{ks.ipynb} details this functionality more completely.

\begin{lstlisting}[style=pythoncode]
# init + train ESN
NR = 1000
esn = orc.forecaster.ESNForecaster(data_dim=3, res_dim=NR, seed=0)
esn, R = orc.forecaster.train_RCForecaster(esn, U_train)
\end{lstlisting}

Once trained, we can perform a forecast in two different ways.

\begin{enumerate}
\tightlist
\item
  Forecast directly from a given reservoir state.
\item
  Forecast from a time-series of "spinup" (also called burn-in) data.
\end{enumerate}

Since we\textquotesingle re forecasting from the last time-step of the
training data, we can use the last entry of \texttt{R} as our reservoir
state from which to forecast or the last 100 timesteps of the training
data as spinup data.

\begin{lstlisting}[style=pythoncode]
# forecast from a given reservoir state
U_pred = esn.forecast(fcast_len=test_timesteps, res_state=R[-1])

# forecast from a time-series of spinup data
# spinup data of last 200 time-steps of training data
spinup_data = U_train[:]
U_pred_IC = esn.forecast_from_IC(
    fcast_len=test_timesteps,
    spinup_data=spinup_data
)

print('Max difference in forecasts:')
print(jnp.max(jnp.abs(U_pred_IC - U_pred)))
\end{lstlisting}

It\textquotesingle s now time to evaluate the accuracy of our forecasts
against the ground truth. Since the Lorenz system is chaotic, we plot
the forecast and ground truth against Lyapunov time. One Lyapunov time
is how long we would expect a perturbation to the initial condition to
grow by a factor of \(e\) and is given by \(1/\lambda_1\) where
\(\lambda_1\) is the largest Lyapunov exponent of the system. We find
that the forecast and ground truth diverge after roughly 8 Lyapunov
times.

\begin{lstlisting}[style=pythoncode]
# plot forecast
LYAP = 0.9
orc.utils.visualization.plot_time_series(
    [U_test, U_pred],
    (t_test - t_test[0]) * LYAP,
    state_var_names=["$u_1$", "$u_2$", "$u_3$"],
    time_series_labels=["True", "Predicted"],
    line_formats=["-", "r--"],
    x_label= r"$\lambda_1 t$",
)
\end{lstlisting}

\pandocbounded{\includegraphics[keepaspectratio]{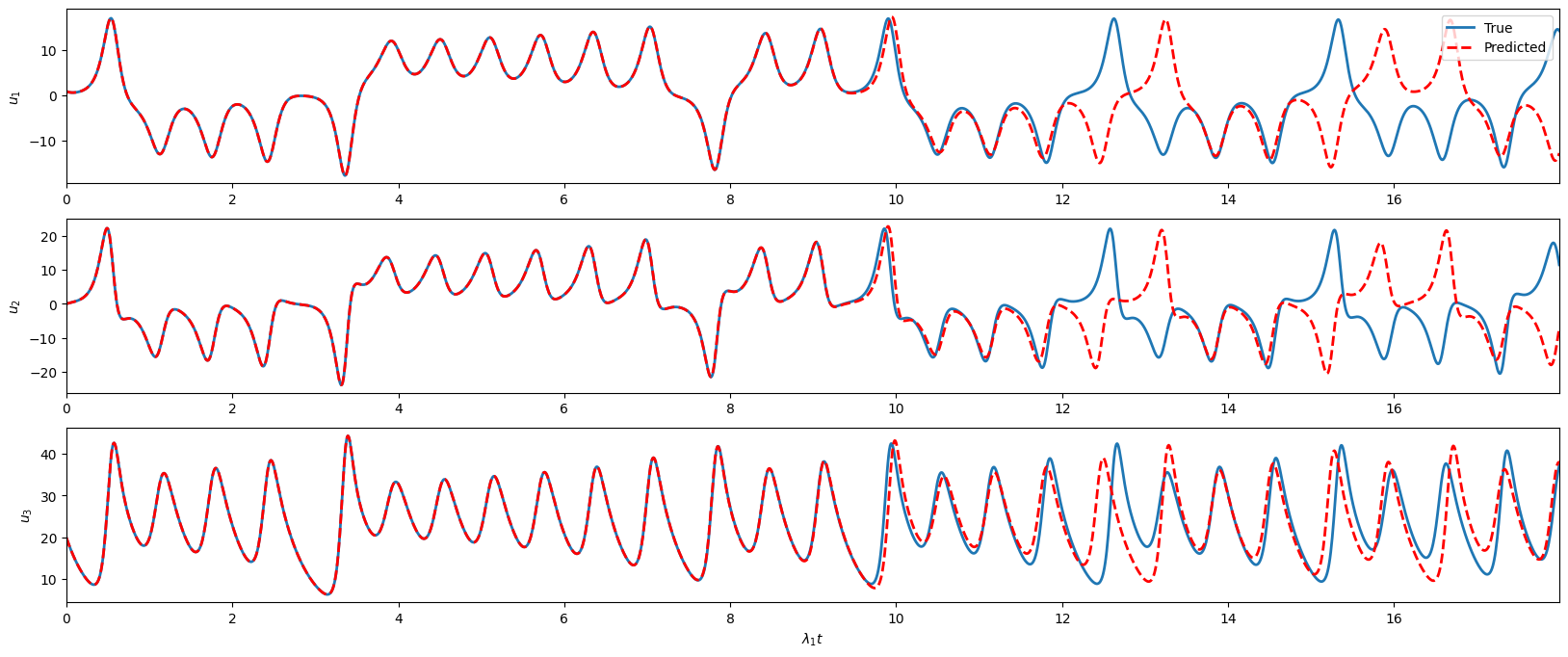}}

\subsection{References}\label{references_lorenz}

{[}1{]} Platt et al., "A systematic exploration of reservoir computing
for forecasting complex spatiotemporal dynamics," \emph{Neural
Networks}, 2022.

\section{Appendix C: Kuramoto-Sivashinsky Forecasting Example}\label{appendix:ks}
This notebook closely resembles Example 1, but instead of forecasting
the Lorenz63 system (which is a low-dimensional, coupled ODE system),
the task at hand is forecasting the Kuramoto Sivashinsky system, a 1D
PDE. This illustrative example also exposes some of the more advanced
functionality of the \texttt{ESNForecaster} model. In particular, a
parallel reservoir scheme is demonstrated and many of the default
hyperparameters of the model are overwritten.

\begin{lstlisting}[style=pythoncode]
import jax
import jax.numpy as jnp
import orc
\end{lstlisting}

\subsection{Generate KS data}\label{generate-ks-data}

Given a domain (\texttt{domain}), spatial discretization (\texttt{Nx}),
temporal discretization (\texttt{dt}), and final time (\texttt{tN}), we
integrate the KS system from a pseudorandom initial condition
(\texttt{u0}). For our specified domain length, we estimate the largest
Lyapunov exponent of the system to be 0.081 {[}1{]}.

\begin{lstlisting}[style=pythoncode]
# integrate
lyap = 0.081
Nx = 128
tN = 1500
domain = (0, 48)
x0 = jnp.linspace(domain[0], domain[1], Nx, endpoint=True)
u0 = jnp.sin((3 / domain[1]) * jnp.pi * x0)
u0 = u0 + jax.random.normal(key=jax.random.key(3), shape=u0.shape)
dt = 0.25
U,t = orc.data.KS_1D(tN, u0=u0, dt=dt, domain=domain, Nx=Nx)
\end{lstlisting}

We visualize the data below using the built in
\texttt{imshow\_1D\_spatiotemp} function.

\begin{lstlisting}[style=pythoncode]
# visualize data
orc.utils.visualization.imshow_1D_spatiotemp(
    U,
    t[-1],
    domain,
    vmax=3.5,
    vmin=-3.5,
    title="KS Data",
    x_label= r"$t$",
)
\end{lstlisting}

\pandocbounded{\includegraphics[keepaspectratio]{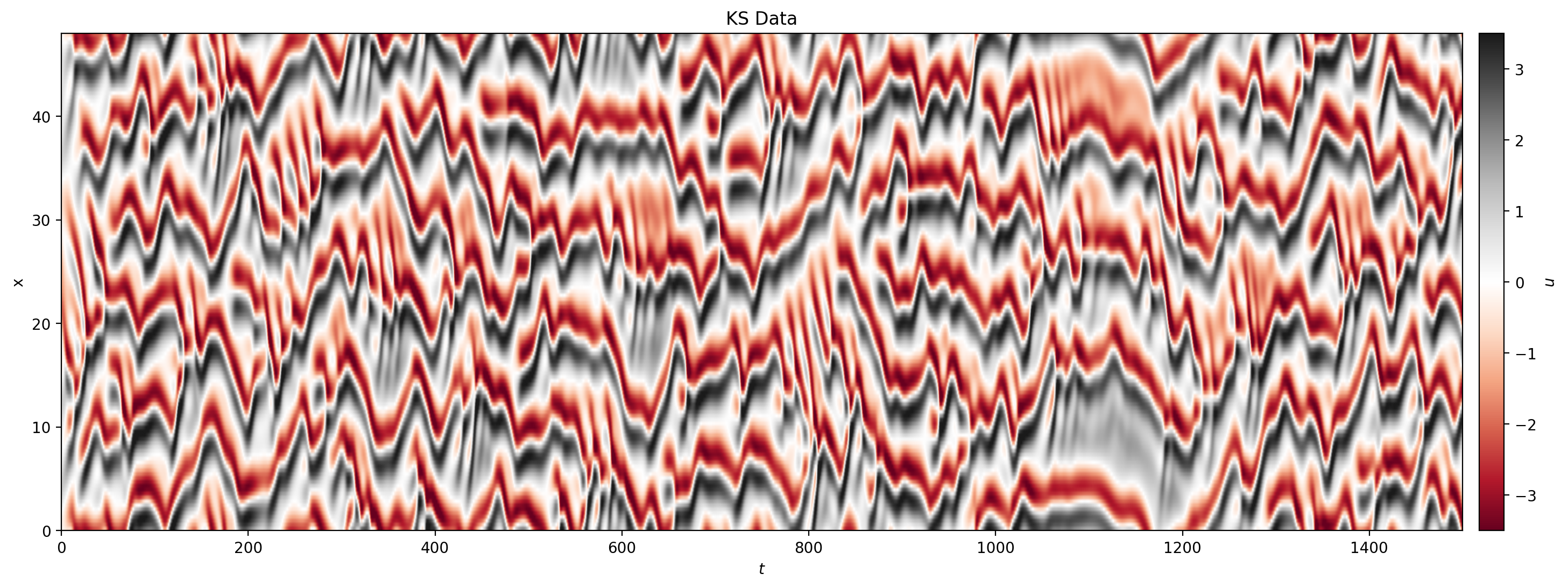}}

As in the Lorenz example, we perform a train/test split. However, in
this case we add \(\approx 3 \%\) Gaussian noise to the training data
because small levels of added noise during training have been shown to
improve the forecasting ability of ESN models {[}2{]}.

\begin{lstlisting}[style=pythoncode]
# train test split
test_perc = 0.2
split_index = int((1 - test_perc) * U.shape[0])
U_train = U[:split_index, :]
train_key = jax.random.key(3)
U_train += jax.random.normal(
    key=train_key,
    shape=U_train.shape
) * jnp.std(U_train) * 0.03
U_test = U[split_index:, :]
t_train = t[:split_index]
t_test = t[split_index:] - t[split_index]
\end{lstlisting}

The work of Pathak et al. {[}3{]} demonstrated that RC based approaches
for forecasting chaotic dynamics could be extended to high-dimensional
cases through the use of a parallel reservoir architecture. Rather than
training a single RC to forecast the entire field, the field is divided
into "chunks" and individual ESNs are trained to forecast a single
chunk. For the architecture to be effective, the input to each reservoir
must include some overlap with adjacent chunks. This parameter is
denoted as \texttt{locality} --- it controls how many grid points from
neighboring chunks each reservoir can see. Choosing \texttt{locality}
relative to the spatial correlation length of the dynamics ensures each
reservoir has enough context to make accurate local predictions.

In addition to providing a convenient interface for training parallel
ESNs by setting the number of chunks \textgreater{} 1 in an
\texttt{ESNForecaster}, the model also accepts other hyperparameter
options as outlined in Platt et al. {[}4{]}. The key hyperparameters are described below:

\begin{itemize}
\tightlist
\item
  \texttt{leak\_rate}: Controls how much of the previous state is
  retained at each step (0 = full memory, 1 = no memory).
\item
  \texttt{embedding\_scaling}: Scales the input-to-reservoir weight
  matrix, controlling input sensitivity.
\item
  \texttt{bias}: Magnitude of the reservoir bias term.
\item
  \texttt{Wr\_spectral\_radius}: Spectral radius of the reservoir weight
  matrix, governing the timescale of reservoir dynamics.
\end{itemize}

\begin{lstlisting}[style=pythoncode]
# esn parameters
res_dim = 1024  # number of reservoir neurons per chunk
chunks = 16     # number of chunks
locality = 8    # locality parameter for reservoir connections
beta = 1e-7

# init esn
esn = orc.forecaster.ESNForecaster(
    data_dim=Nx,
    res_dim=res_dim,
    seed=2,
    chunks=chunks,
    locality=locality,
    leak_rate=0.534,
    embedding_scaling=0.005,
    bias=1.915,
    Wr_spectral_radius=0.7,
    )
\end{lstlisting}

The instantiated \texttt{ESNForecaster} is trained as in the Lorenz
example.

\begin{lstlisting}[style=pythoncode]
esn, R = orc.forecaster.train_RCForecaster(
    esn,
    U_train,
    beta=beta
)
\end{lstlisting}

Forecasting with the parallel reservoir ESN is also performed exactly as
in the single reservoir case.

\begin{lstlisting}[style=pythoncode]
U_fcast = esn.forecast(U_test.shape[0], R[-1])
\end{lstlisting}

We now plot the forecast of the trained ESN against Lyapunov time\ldots

\begin{lstlisting}[style=pythoncode]
orc.utils.visualization.imshow_1D_spatiotemp(
    U_fcast,
    t_test[-1] * lyap,
    domain,
    vmax=3.5,
    vmin=-3.5,
    title="KS Forecast",
    x_label= r"$\lambda _1 t$"
)
\end{lstlisting}

\pandocbounded{\includegraphics[keepaspectratio]{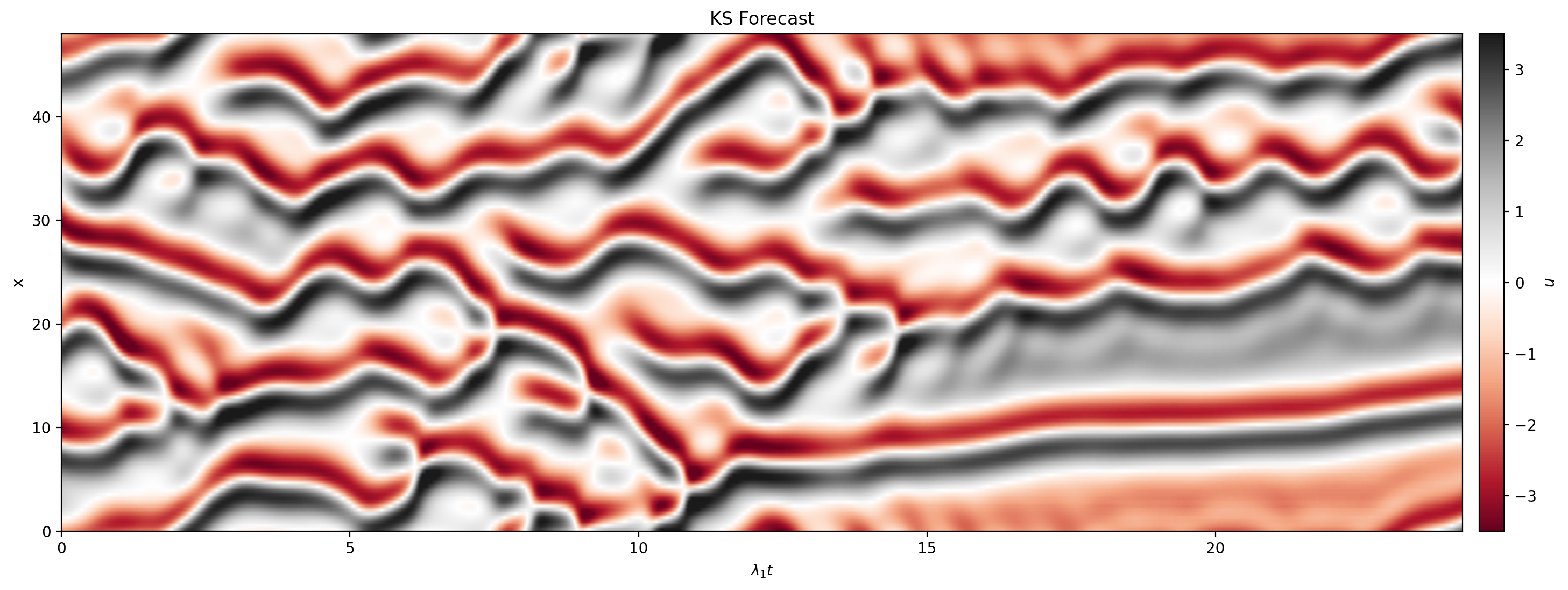}}

\ldots{} as well as the deviation from the ground truth test data.

\begin{lstlisting}[style=pythoncode]
orc.utils.visualization.imshow_1D_spatiotemp(
    U_test - U_fcast,
    t_test[-1] * lyap,
    domain,
    vmax=3.5,
    vmin=-3.5,
    title="KS Forecast Error",
    x_label= r"$\lambda _1 t$"
)
\end{lstlisting}

\pandocbounded{\includegraphics[keepaspectratio]{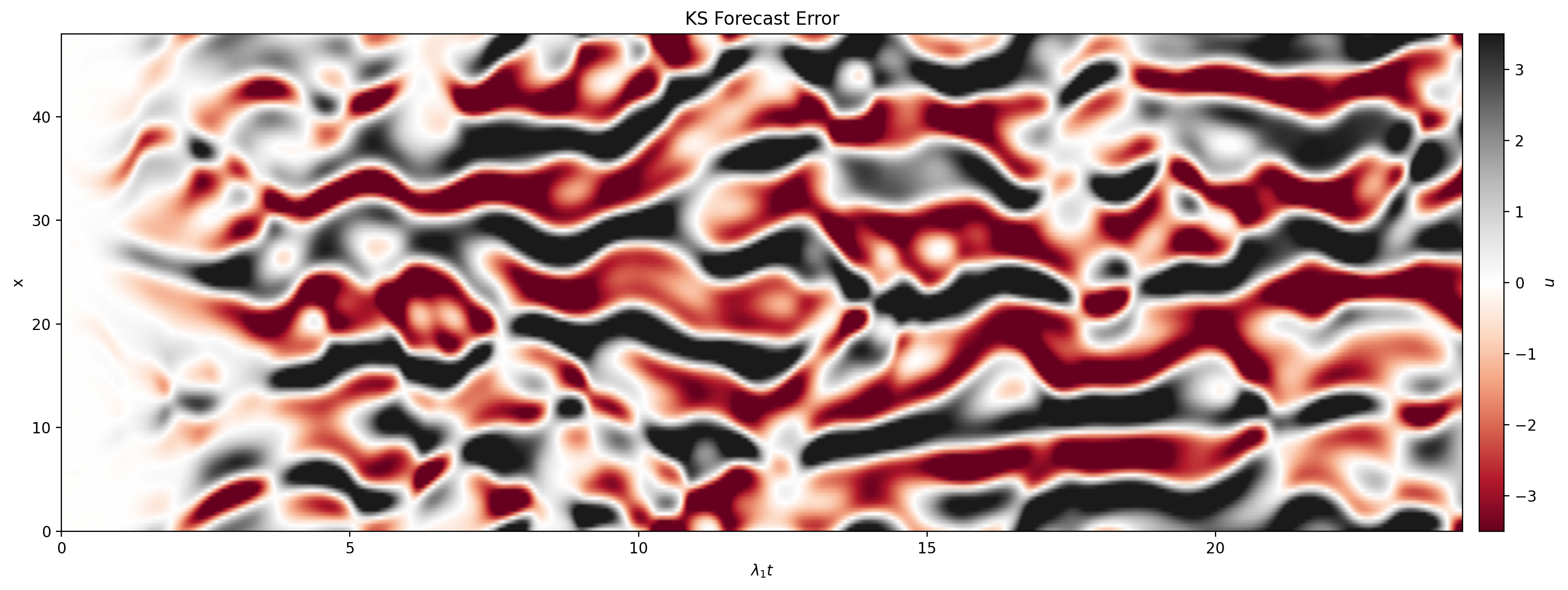}}

Forecasting performance can be improved (at the cost of computation
time) by increasing both the number of chunks and the reservoir
dimensions.

\subsection{References}\label{references_ks}

{[}1{]} Edson et al., "Lypaunov exponents of the Kuramoto Sivashinsky
PDE," \emph{The Anziam Journal}, 2019.

{[}2{]} Vlachas et al., "Backpropagation Algorithms and Reservoir
Computing in Recurrent Neural Networks for the Forecasting of Complex
Spatiotemporal Dynamics," \emph{Neural Networks}, 2020.

{[}3{]} Pathak et al., "Model-Free Prediction of Large Spatiotemporally
Chaotic Systems from Data: A Reservoir Computing Approach,"
\emph{Physical Review Letters}, 2018.

{[}4{]} Platt et al., "A systematic exploration of reservoir computing
for forecasting complex spatiotemporal dynamics," \emph{Neural
Networks}, 2022.

\section{Appendix D: Classification Example}\label{appendix:classification}
This notebook demonstrates how to use ORC\textquotesingle s classifier
module to classify time-series data using an Echo State Network (ESN).
We generate synthetic time-series belonging to distinct classes, train
an \texttt{ESNClassifier}, and evaluate its performance. This is a
common task in areas like activity recognition, medical signal analysis,
and sensor data classification.

As with other ORC modules, only \texttt{jax}, \texttt{orc}, and
\texttt{equinox} are needed. We also import \texttt{matplotlib} for
visualization.

\begin{lstlisting}[style=pythoncode]
import jax
import jax.numpy as jnp
import equinox as eqx
import matplotlib.pyplot as plt
import orc
\end{lstlisting}

\subsection{Generate Synthetic Classification
Data}\label{generate-synthetic-classification-data}

We create a simple 3-class classification problem where each class is a
sinusoidal time-series with a distinct frequency. A small amount of
noise is added to each sample. This mimics the kind of structure found
in real-world time-series classification tasks, where different classes
exhibit different temporal patterns.

\begin{lstlisting}[style=pythoncode]
# Dataset parameters
data_dim = 3
n_classes = 3
seq_len = 200
n_train_per_class = 20
n_test_per_class = 10

key = jax.random.PRNGKey(0)

def generate_samples(key, n_per_class):
    seqs, labels = [], []
    t = jnp.linspace(0, 4 * jnp.pi, seq_len).reshape(-1, 1)
    for class_idx in range(n_classes):
        freq = (class_idx + 1) * 0.5
        for _ in range(n_per_class):
            key, subkey = jax.random.split(key)
            noise = 0.05 * jax.random.normal(subkey, (seq_len, data_dim))
            seq = jnp.sin(freq * t * jnp.arange(1, data_dim + 1)) + noise
            seqs.append(seq)
            labels.append(class_idx)
    return jnp.stack(seqs), jnp.array(labels, dtype=jnp.int32), key

train_seqs, train_labels, key = generate_samples(key, n_train_per_class)
test_seqs, test_labels, key = generate_samples(key, n_test_per_class)

print(f"Training set: {train_seqs.shape}  (n_samples, seq_len, data_dim)")
print(f"Test set:     {test_seqs.shape}")
\end{lstlisting}

Let\textquotesingle s visualize one example from each class to see how
the temporal patterns differ.

\begin{lstlisting}[style=pythoncode]
fig, axes = plt.subplots(1, n_classes, figsize=(14, 3), sharey=True)
class_names = ["Low freq", "Medium freq", "High freq"]
for i in range(n_classes):
    idx = i * n_train_per_class
    axes[i].plot(train_seqs[idx, :, 0])
    axes[i].set_title(f"Class {i}: {class_names[i]}")
    axes[i].set_xlabel("Time step")
axes[0].set_ylabel("Amplitude (channel 0)")
plt.tight_layout()
plt.show()
\end{lstlisting}

\pandocbounded{\includegraphics[keepaspectratio]{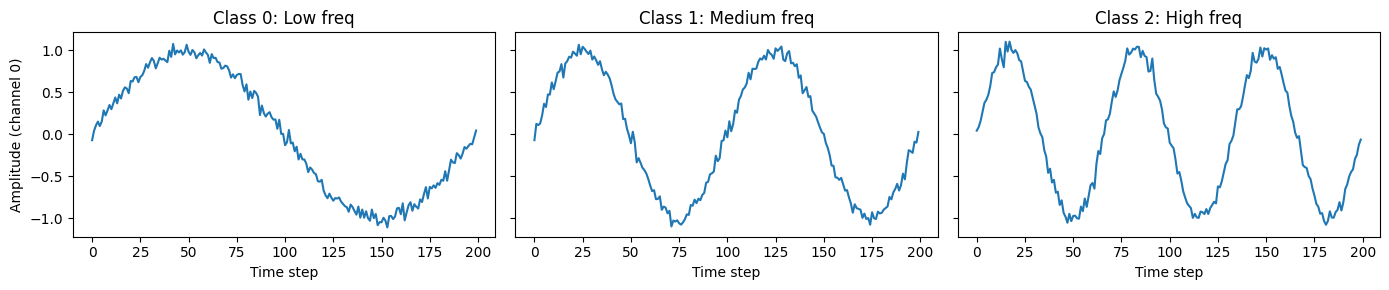}}

\subsection{Initialize the ESN
Classifier}\label{initialize-the-esn-classifier}

The \texttt{ESNClassifier} follows the same Embedding \(\to\) Driver
\(\to\) Readout architecture as other ORC models. To instantiate a
classifier, we specify:

\begin{itemize}
\tightlist
\item
  \texttt{data\_dim}: the number of input channels per time-step
\item
  \texttt{n\_classes}: the number of classification classes
\item
  \texttt{res\_dim}: the dimension of the reservoir (latent state)
\end{itemize}

The \texttt{state\_repr} parameter controls how reservoir states are
summarized into a feature vector for classification:

\begin{itemize}
\tightlist
\item
  \texttt{"final"} (default): uses only the last reservoir state
\item
  \texttt{"mean"}: averages all reservoir states (after an optional
  spinup period)
\end{itemize}

\begin{lstlisting}[style=pythoncode]
res_dim = 500

classifier = orc.classifier.ESNClassifier(
    data_dim=data_dim,
    n_classes=n_classes,
    res_dim=res_dim,
    seed=42,
)

print(f"Reservoir dimension: {classifier.res_dim}")
print(f"Input dimension:     {classifier.in_dim}")
print(f"Output dimension:    {classifier.out_dim}")
print(f"State representation: {classifier.state_repr}")
\end{lstlisting}

\subsection{Train the Classifier}\label{train-the-classifier}

Training is performed by \texttt{orc.classifier.train\_ESNClassifier}.
Under the hood, this:

\begin{enumerate}
\tightlist
\item
  Forces each training sequence through the reservoir (in parallel via
  \texttt{jax.vmap})
\item
  Extracts a feature vector per sequence (final state or mean state)
\item
  Solves ridge regression from features to one-hot encoded class labels
\item
  Updates the readout weights
\end{enumerate}

The \texttt{beta} parameter controls Tikhonov regularization.

\begin{lstlisting}[style=pythoncode]
trained_classifier = orc.classifier.train_RCClassifier(
    classifier,
    train_seqs=train_seqs,
    labels=train_labels,
    beta=1e-6,
)

print("Training complete.")
\end{lstlisting}

\subsection{Evaluate on Training Data}\label{evaluate-on-training-data}

We can classify individual sequences using \texttt{classify}, which by
default zero-initializes the reservoir state and returns a vector of
class probabilities (via softmax). The predicted class is the one with
the highest probability.

\begin{lstlisting}[style=pythoncode]
probs = jax.vmap(trained_classifier.classify)(train_seqs)

train_preds = jnp.argmax(probs, axis=1)
train_acc = jnp.mean(train_preds == train_labels)
print(f"Training accuracy: {train_acc:.1%}")
\end{lstlisting}

\subsection{Evaluate on Test Data}\label{evaluate-on-test-data}

The real test of the classifier is on unseen data. We classify each test
sequence and compute the accuracy.

\begin{lstlisting}[style=pythoncode]
probs = jax.vmap(trained_classifier.classify)(test_seqs)

test_preds = jnp.argmax(probs, axis=1)
test_acc = jnp.mean(test_preds == test_labels)
print(f"Test accuracy: {test_acc:.1%}")
\end{lstlisting}

Let\textquotesingle s look at the predicted probabilities for a few test
samples to see how confident the classifier is.

\begin{lstlisting}[style=pythoncode]
fig, axes = plt.subplots(1, 3, figsize=(12, 3))
for i, class_idx in enumerate(range(n_classes)):
    sample_idx = class_idx * n_test_per_class
    probs = trained_classifier.classify(test_seqs[sample_idx])
    axes[i].bar(
        range(n_classes),
        probs,
        color=["#4C72B0", "#55A868", "#C44E52"]
    )
    axes[i].set_xticks(range(n_classes))
    axes[i].set_xticklabels(class_names)
    axes[i].set_ylim(0, 1)
    axes[i].set_title(f"True: {class_names[class_idx]}")
    axes[i].set_ylabel("Probability" if i == 0 else "")
plt.suptitle("Predicted Class Probabilities", y=1.02)
plt.tight_layout()
plt.show()
\end{lstlisting}

\pandocbounded{\includegraphics[keepaspectratio]{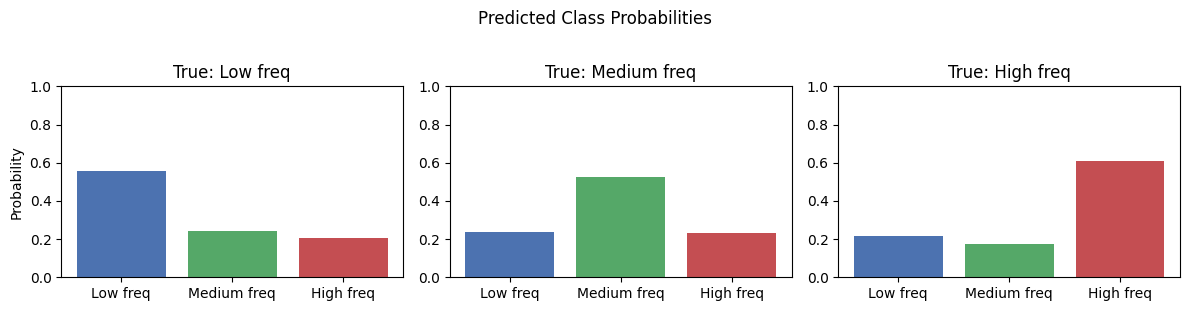}}

\subsection{Using Mean State
Representation}\label{using-mean-state-representation}

Instead of using only the final reservoir state for classification, we
can average the reservoir states over the sequence. This is often more
robust when the discriminative information is spread throughout the
time-series rather than concentrated at the end --- for example, when
class identity is determined by frequency content or statistical
properties of the entire signal. The \texttt{spinup} parameter discards
early transient states before averaging, since the reservoir needs time
to synchronize with the input.

\begin{lstlisting}[style=pythoncode]
classifier_mean = orc.classifier.ESNClassifier(
    data_dim=data_dim,
    n_classes=n_classes,
    res_dim=res_dim,
    seed=42,
    state_repr="mean",
)

trained_mean = orc.classifier.train_RCClassifier(
    classifier_mean,
    train_seqs=train_seqs,
    labels=train_labels,
    spinup=20,
    beta=1e-6,
)

in_seq = test_seqs
res_state = None
spinup = 20
probs = eqx.filter_vmap(trained_mean.classify)(
    in_seq,
    res_state,
    spinup
    )

mean_preds = jnp.argmax(probs, axis=1)
mean_acc = jnp.mean(mean_preds == test_labels)
print(f"Test accuracy (mean state): {mean_acc:.1%}")
\end{lstlisting}

\subsection{Summary}\label{summary_class}

This notebook demonstrated the core classification workflow in ORC:

\begin{enumerate}
\tightlist
\item
  \textbf{Initialize} an \texttt{ESNClassifier} with
  \texttt{orc.classifier.ESNClassifier}
\item
  \textbf{Train} with \texttt{orc.classifier.train\_ESNClassifier},
  which uses ridge regression for a closed-form solution
\item
  \textbf{Classify} new sequences with \texttt{classify} (reservoir
  state defaults to zero if not provided)
\end{enumerate}

Key parameters to tune:

\begin{itemize}
\tightlist
\item
  \texttt{res\_dim}: larger reservoirs can capture more complex dynamics
\item
  \texttt{beta}: regularization strength (increase if overfitting)
\item
  \texttt{state\_repr}: \texttt{"final"} vs \texttt{"mean"} depending on
  where class-discriminative information lies in the sequence
\item
  \texttt{spinup}: number of transient states to discard when using
  \texttt{state\_repr="mean"}
\end{itemize}

\section{Appendix E: Control Example}\label{appendix:control}
This notebook demonstrates using an ESN as a learned model for
model-predictive control (MPC). The idea is to train an ESN to predict
how a system responds to control inputs, then use that learned model to
optimize control sequences that drive the system toward a desired
reference trajectory.

The controller minimizes a cost function over a finite forecast horizon:

\[L(\mathbf{u}) = \alpha_1 \|\mathbf{y}(\mathbf{u}) - \mathbf{y}_{\text{ref}}\|^2 + \alpha_2 \|\mathbf{u}\|^2 + \alpha_3 \|\Delta \mathbf{u}\|^2\]

where \(\mathbf{y}(\mathbf{u})\) is the ESN\textquotesingle s predicted
output given control sequence \(\mathbf{u}\),
\(\mathbf{y}_{\text{ref}}\) is the reference trajectory, and
\(\Delta \mathbf{u}\) penalizes non-smooth control. Optimization is
performed via BFGS.

\begin{lstlisting}[style=pythoncode]
import orc
import jax
import jax.numpy as jnp
import matplotlib.pyplot as plt
\end{lstlisting}

\subsection{Plant: Two-Mass Spring-Damper
System}\label{plant-two-mass-spring-damper-system}

Our test system is a wall-coupled two-mass spring-damper:

\begin{verbatim}
Wall ---[k1, c1]--- m1 ---[k2, c2]--- m2
\end{verbatim}

The state vector is \([x_1, x_2, v_1, v_2]\) (positions and velocities
of both masses). A force applied to \(m_1\) is the control input, and we
observe the positions \([x_1, x_2]\).

\begin{lstlisting}[style=pythoncode]
class TwoMassSpringSystem:
    """
    Two-mass spring-damper system connected to a wall.

    Wall ---[k1, c1]--- m1 ---[k2, c2]--- m2

    State vector: [x1, x2, v1, v2] (positions and velocities)
    Input: Force applied to mass 1
    Output: Positions of both masses [x1, x2]
    """

    def __init__(
        self,
        m1=1.0,
        m2=1.0,
        k1=1.0,
        k2=1.0,
        c1=0.0,
        c2=0.0,
        dt=0.01
    ):
        self.m1 = m1
        self.m2 = m2
        self.k1 = k1
        self.k2 = k2
        self.c1 = c1
        self.c2 = c2
        self.dt = dt

        # Continuous-time state matrix with damping
        # State: [x1, x2, v1, v2]
        A_cont = jnp.array([
            [0, 0, 1, 0],
            [0, 0, 0, 1],
            [-(k1 + k2) / m1, k2 / m1, -(c1 + c2) / m1, c2 / m1],
            [k2 / m2, -k2 / m2, c2 / m2, -c2 / m2]
        ])

        # Continuous-time input matrix (force on m1)
        B_cont = jnp.array([[0], [0], [1 / m1], [0]])

        # Discretize using Euler method
        self.A = jnp.eye(4) + dt * A_cont
        self.B = dt * B_cont

        # Output matrix: observe positions [x1, x2]
        self.C = jnp.array([
            [1, 0, 0, 0],
            [0, 1, 0, 0]
        ])
        self.D = jnp.zeros((2, 1))

        self.state_dim = 4
        self.output_dim = 2

    def step(self, x, u):
        """Single timestep."""
        u = jnp.atleast_1d(u).reshape(-1, 1)
        x_next = self.A @ x + self.B @ u
        y = self.C @ x_next + self.D @ u
        return x_next, y.flatten()

    def simulate(self, u_sequence, x0=None):
        """Simulate over input sequence."""
        T = len(u_sequence)
        x = x0 if x0 is not None else jnp.zeros((self.state_dim, 1))

        states = []
        outputs = []

        for t in range(T):
            x, y = self.step(x, u_sequence[t])
            states.append(x.flatten())
            outputs.append(y)

        return jnp.stack(states), jnp.stack(outputs)
\end{lstlisting}

\subsection{Generate Training Data}\label{generate-training-data}

To train the ESN controller, we need input-output pairs that explore the
system\textquotesingle s behavior. We apply a random piecewise-constant
force to \(m_1\) and record the resulting positions. This gives the ESN
a diverse set of dynamics to learn from.

\begin{lstlisting}[style=pythoncode]
system_damped = TwoMassSpringSystem(
    m1=1.0,
    m2=1.0,
    k1=2.0,
    k2=1.0,
    c1=0.4,
    c2=0.4,
    dt=0.1
)

# Impulse input
T = 1000
t = jnp.arange(T) * 0.1
u = jax.random.uniform(
    jax.random.key(0),
    shape=(100,),
    minval=-1,
    maxval=1
)
u = jnp.repeat(u, 10).reshape(-1, 1)
print(u.shape)

states_damped, outputs_damped = system_damped.simulate(u)

plt.figure(figsize=(10, 4))
plt.plot(t, outputs_damped[:, 0], 'b-', label='x1 damped')
plt.plot(t, outputs_damped[:, 1], 'r-', label='x2 damped')
plt.xlabel('Time')
plt.ylabel('Position')
plt.legend()
plt.title('Training Trajectory')
plt.show()
\end{lstlisting}

\pandocbounded{\includegraphics[keepaspectratio]{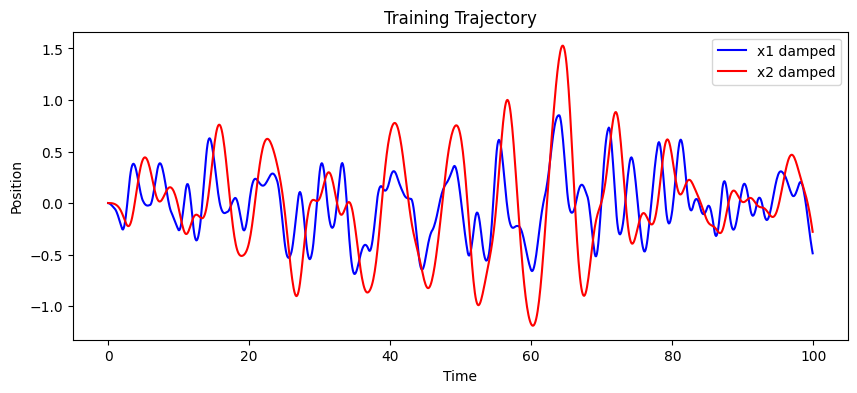}}

\subsection{Train the ESN Controller}\label{train-the-esn-controller}

\texttt{ESNController} takes both the observed outputs and control
inputs as input to the reservoir embedding (concatenated). Training uses
ridge regression, just like the forecasting models.

\begin{lstlisting}[style=pythoncode]
control_model = orc.control.ESNController(
    data_dim=2,
    control_dim=1,
    res_dim=1000
)
control_model, R = orc.control.train_RCController(
    control_model,
    outputs_damped[:-1],
    u[1:]
)
\end{lstlisting}

\subsection{Closed-Loop Control}\label{closed-loop-control}

We now run the controller in a receding-horizon loop. At each step:

\begin{enumerate}
\tightlist
\item
  \texttt{compute\_control()} optimizes a 20-step control sequence to
  track the reference trajectory (all zeros --- we want to bring the
  system to rest).
\item
  Only the first control input is applied to the real system.
\item
  The reservoir state is updated with the observed output and applied
  control via \texttt{force()}.
\item
  The process repeats from the new state.
\end{enumerate}

This is the standard model-predictive control (MPC) pattern, with the
ESN serving as the predictive model.

\begin{lstlisting}[style=pythoncode]
ref_traj = jnp.zeros((500, 2))
r0 = R[-1]
state, output = states_damped[-1:], outputs_damped[-1:]
state_list = []
control_list = []

for i in range(250):
    comp_control = control_model.compute_control(
        jnp.zeros((20, 1)), res_state=r0, ref_traj=ref_traj[i:i+20]
    )

    r0 = control_model.force(output, comp_control[0:1,0:1], r0)[0]

    state, output = system_damped.simulate(
        comp_control[0:1,0:1],
        state.reshape(4,-1)
    )
    state_list.append(state)
    control_list.append(comp_control[0,0])


state_list = jnp.array(state_list)

control_list = jnp.array(control_list)
\end{lstlisting}

\subsection{Compare Controlled vs.
Uncontrolled}\label{compare-controlled-vs-uncontrolled}

For comparison, we simulate the system from the same initial state with
zero input (no control). The controlled system should converge to the
reference (zero displacement), while the uncontrolled system oscillates
freely.

\begin{lstlisting}[style=pythoncode]
uncontrolled_states, uncontrolled_outputs = system_damped.simulate(
    jnp.zeros((250,1)),
    states_damped[-1:].reshape(4,-1)
)
\end{lstlisting}

\begin{lstlisting}[style=pythoncode]
t = jnp.arange(1, 251) * 0.1
plt.figure(figsize=(10, 4))
plt.plot(t, state_list[:, 0, [0,1]], 'b-', label='Controlled')
plt.plot(t, uncontrolled_outputs, 'r-', label='Uncontrolled')
plt.xlabel('Time')
plt.ylabel('Position')
plt.legend()
plt.title('Effect of Control')
plt.show()
\end{lstlisting}

\pandocbounded{\includegraphics[keepaspectratio]{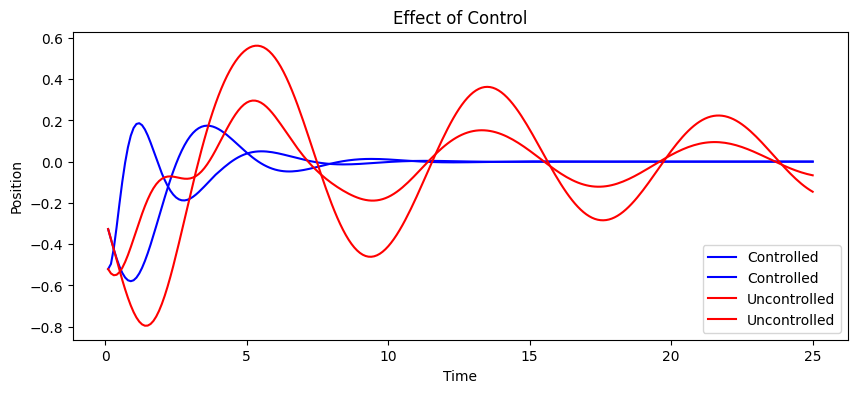}}

\section{Appendix F: Continuous-Time Reservoir Computing}\label{appendix:continuous-rc}
Standard ESNs operate in discrete time --- the reservoir state updates
via a map at each timestep. A \textbf{continuous-time ESN} (CESN)
instead evolves the reservoir state as an ODE:

\[\frac{d\mathbf{r}}{dt} = \tau\left(-\mathbf{r} + \tanh\left(W_r \mathbf{r} + W_{in} \mathbf{u}(t) + \mathbf{b}\right)\right)\]

where \(\tau\) is a time constant controlling how quickly the reservoir
responds to input. During training, the input \(\mathbf{u}(t)\) is
interpolated continuously (cubic Hermite) and the ODE is solved with an
adaptive solver via
\href{https://github.com/patrick-kidger/diffrax}{Diffrax}.

This formulation is useful when the underlying dynamics are naturally
continuous, when data is irregularly sampled, or when you want explicit
control over integration accuracy.

\begin{lstlisting}[style=pythoncode]
import jax
import jax.numpy as jnp
import diffrax
import equinox as eqx
from jaxtyping import Array, Float

jax.config.update("jax_enable_x64", True)
import orc
import orc.utils.visualization as vis
\end{lstlisting}

\subsection{Data}\label{data}

\begin{lstlisting}[style=pythoncode]
tN = 100.0
dt = 0.02
test_perc = 0.2
U, t = orc.data.lorenz63(tN=tN, dt=dt)
vis.plot_time_series(U, t)

# train test split
split_idx = int((1 - test_perc) * U.shape[0])
U_train = U[:split_idx]
U_test = U[split_idx:]
t_train = t[:split_idx]
t_test = jnp.arange(U_test.shape[0]) * dt

print(f"Train shape: {U_train.shape}, Test shape: {U_test.shape}")
\end{lstlisting}

\pandocbounded{\includegraphics[keepaspectratio]{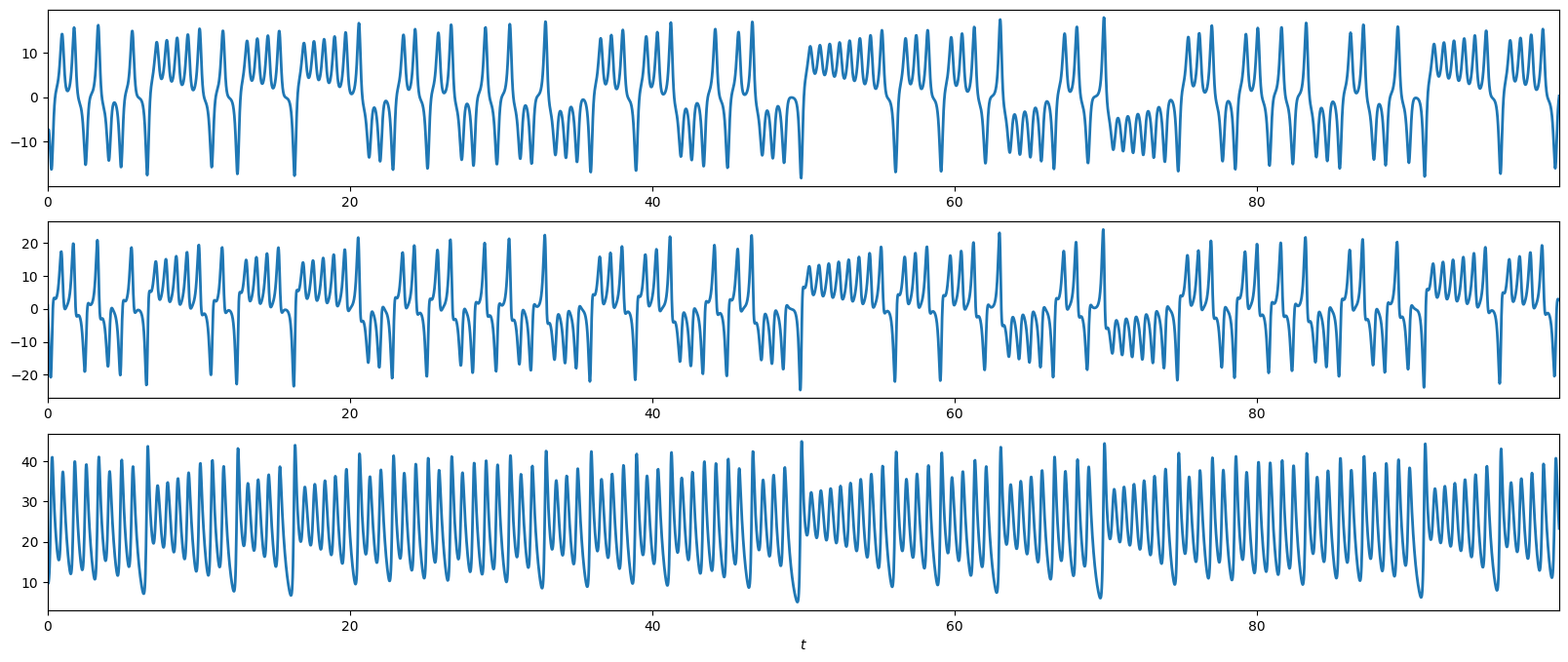}}

\subsection{Train a Continuous ESN}\label{train-a-continuous-esn}

We create a \texttt{CESNForecaster} with \texttt{time\_const=40.0}. The
time constant \(\tau\) sets the reservoir\textquotesingle s intrinsic
timescale relative to the data --- larger values produce slower, more
stable dynamics. Choosing \(\tau\) appropriately for your
system\textquotesingle s timescale is important for good performance.

By default, \texttt{CESNForecaster} uses a 5th-order Runge-Kutta solver
(\texttt{Tsit5}) with adaptive step-size control
(\texttt{PIDController}). This balances accuracy and computational cost
automatically.

\begin{lstlisting}[style=pythoncode]
esn = orc.forecaster.CESNForecaster(
    data_dim=U_train.shape[1],
    res_dim=1000,
    time_const=40.0
)
esn, R = orc.forecaster.train_RCForecaster(
    model=esn,
    train_seq=U_train,
    ts=t_train
)
\end{lstlisting}

\begin{lstlisting}[style=pythoncode]
U_pred = esn.forecast(ts=t_test, res_state=R[-1])
vis.plot_time_series(
    [U_test, U_pred],
    t_test,
    line_formats=["-", "r--"],
    time_series_labels=["True", "Predicted"],
)
\end{lstlisting}

\pandocbounded{\includegraphics[keepaspectratio]{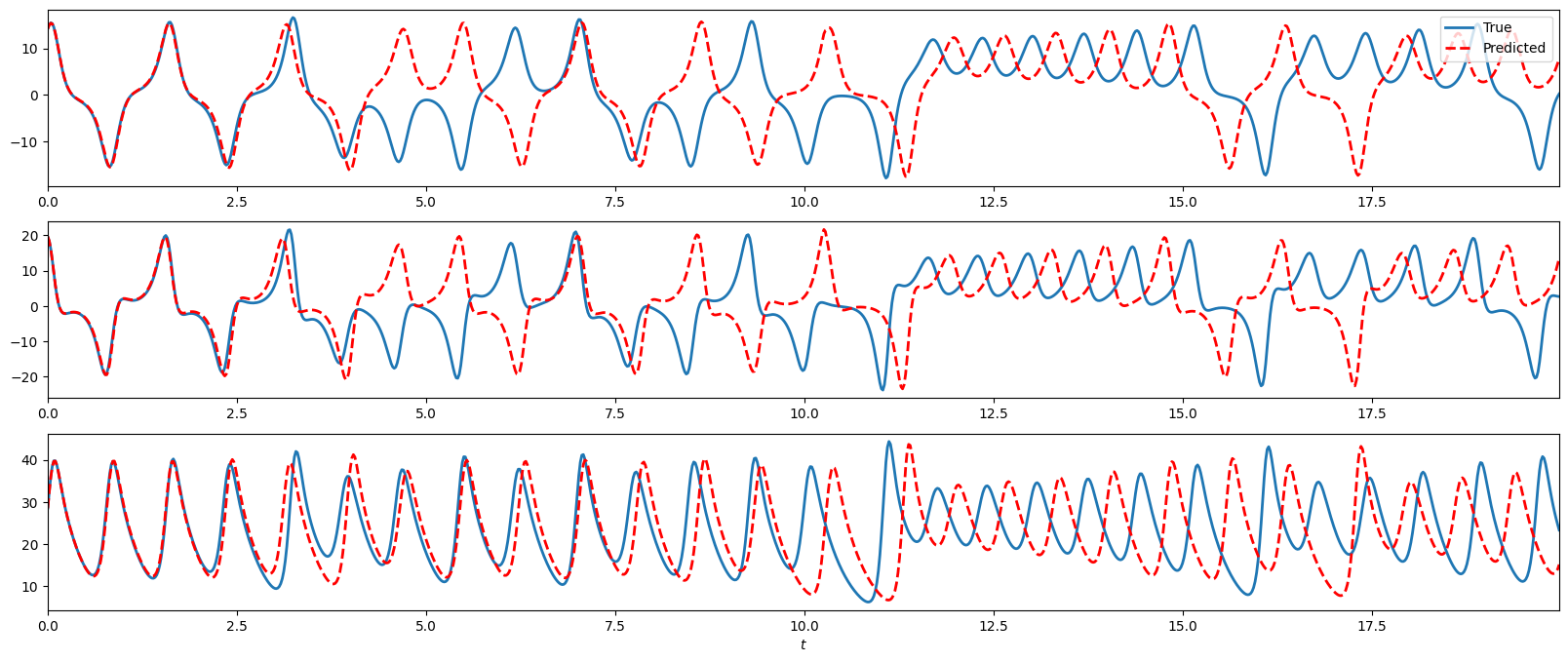}}

\subsection{Adjust ODE Solver and Step-Size
Controller}\label{adjust-ode-solver-and-step-size-controller}

Since \texttt{CESNForecaster} delegates integration to Diffrax, the user
can swap in any compatible solver or step-size controller. Here we
switch to a fixed-step Euler method (\texttt{diffrax.Euler} +
\texttt{diffrax.ConstantStepSize}).

\begin{lstlisting}[style=pythoncode]
solver = diffrax.Euler()
stepsize_controller = diffrax.ConstantStepSize()
esn = orc.forecaster.CESNForecaster(
    data_dim=U_train.shape[1],
    res_dim=400,
    time_const=50.0,
    solver=solver,
    stepsize_controller=stepsize_controller
)
esn, R = orc.forecaster.train_CESNForecaster(
    model=esn,
    train_seq=U_train,
    t_train=t_train
)
\end{lstlisting}

\begin{lstlisting}[style=pythoncode]
U_pred = esn.forecast(ts=t_test, res_state=R[-1])
vis.plot_time_series(
    [U_test, U_pred],
    t_test,
    line_formats=["-", "r--"],
    time_series_labels=["True", "Predicted"],
)
\end{lstlisting}

\pandocbounded{\includegraphics[keepaspectratio]{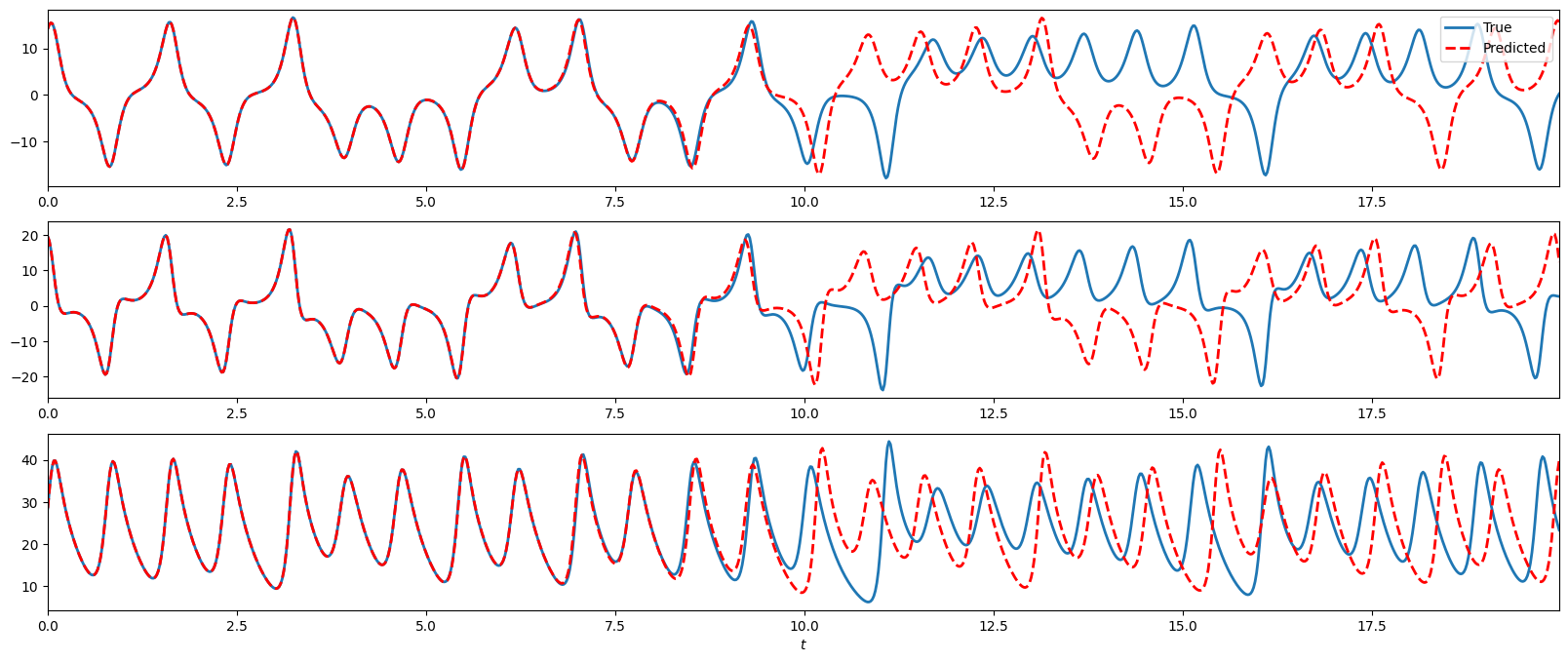}}

\subsection{Forecast from Initial
Conditions}\label{forecast-from-initial-conditions}

When forecasting a new trajectory (not continuing from the training
data), the reservoir must first be "spun up" --- driven with a short
segment of real data to synchronize its internal state with the
system\textquotesingle s current dynamics. Without spinup, the reservoir
starts from a zero state that doesn\textquotesingle t reflect the true
system, leading to poor initial predictions.

\begin{lstlisting}[style=pythoncode]
# Some new data
tN = 20.0
dt = 0.02
eps = 50  # amount of spinup data

U, t = orc.data.lorenz63(tN=tN, dt=dt, u0 = [10,-1,10])
U_spinup = U[:eps]
U_test2 = U[eps:]
t_test2 = jnp.arange(U_test2.shape[0]) * dt
vis.plot_time_series(U_test2, t_test2)
\end{lstlisting}

\pandocbounded{\includegraphics[keepaspectratio]{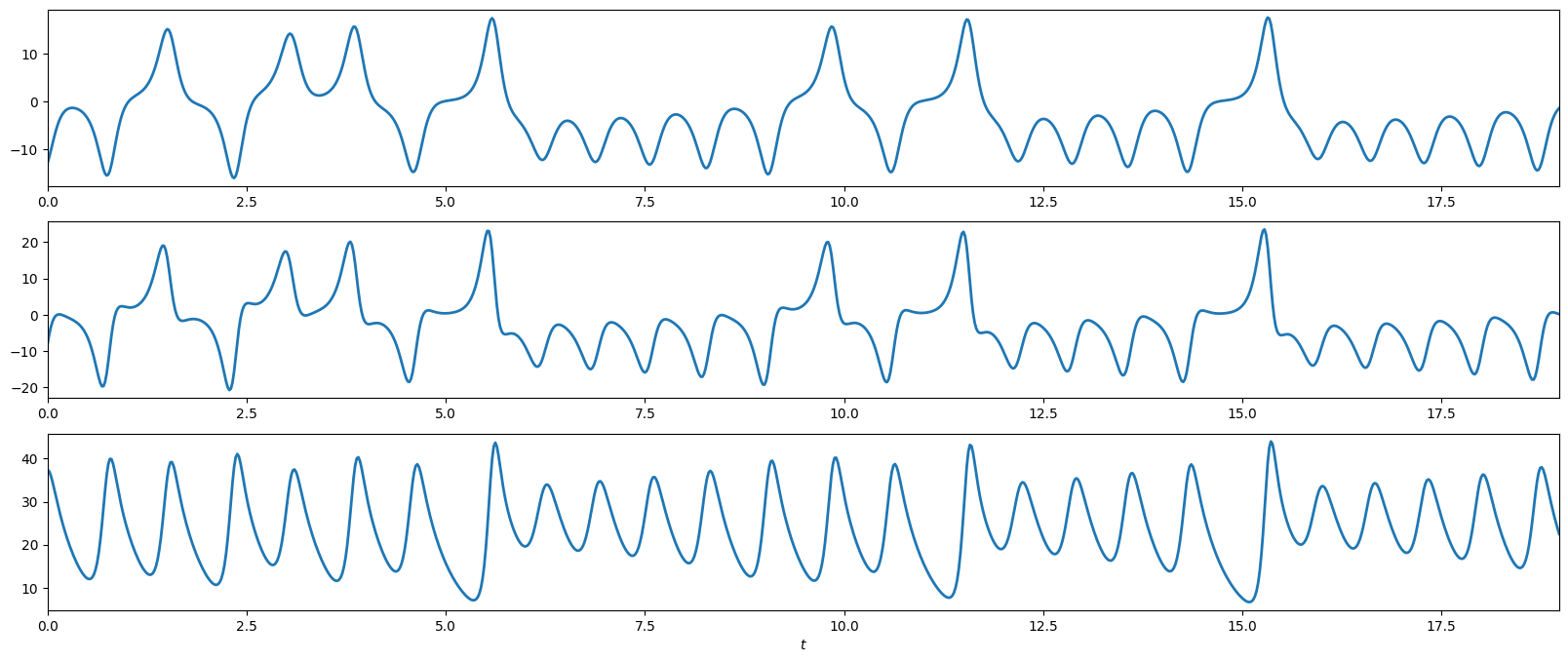}}

\begin{lstlisting}[style=pythoncode]
U_pred2 = esn.forecast_from_IC(ts=t_test2, spinup_data=U_spinup)
vis.plot_time_series(
    [U_test2, U_pred2],
    t_test2,
    line_formats=["-", "r--"],
    time_series_labels=["True", "Predicted"],
)
\end{lstlisting}

\pandocbounded{\includegraphics[keepaspectratio]{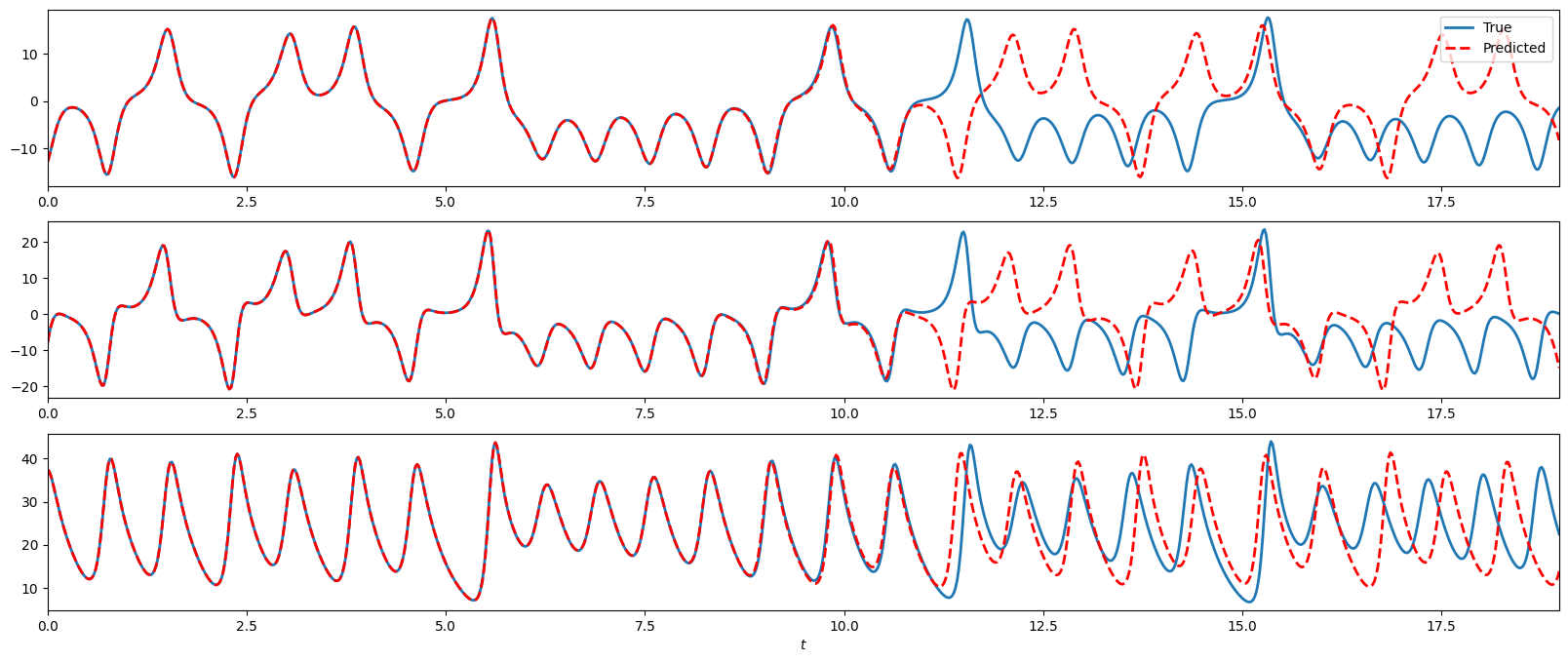}}

\section{Appendix G: Data Library}\label{appendix:data-library}
The \texttt{orc.data} module provides a collection of ODE and PDE
benchmark systems commonly used to evaluate reservoir computing methods.
Each function returns a solution array \texttt{U} and time vector
\texttt{t}.

\begin{lstlisting}[style=pythoncode]
import functools
import time

import jax
import jax.numpy as jnp
import jax.random
import diffrax
import numpy as np
import matplotlib.pyplot as plt

import orc.data
import orc.utils.visualization as vis

jax.config.update("jax_enable_x64", True)
\end{lstlisting}

\subsection{ODE Systems}\label{ode-systems}

\textbf{Lorenz63} --- 3D chaotic attractor and the canonical RC
benchmark. Lyapunov time \(\approx\) 1.1 time units.

\begin{lstlisting}[style=pythoncode]
u,t = orc.data.lorenz63(tN = 20, dt = 0.01)
vis.plot_time_series(u,t, title="Lorenz63 Time Series")
\end{lstlisting}

\pandocbounded{\includegraphics[keepaspectratio]{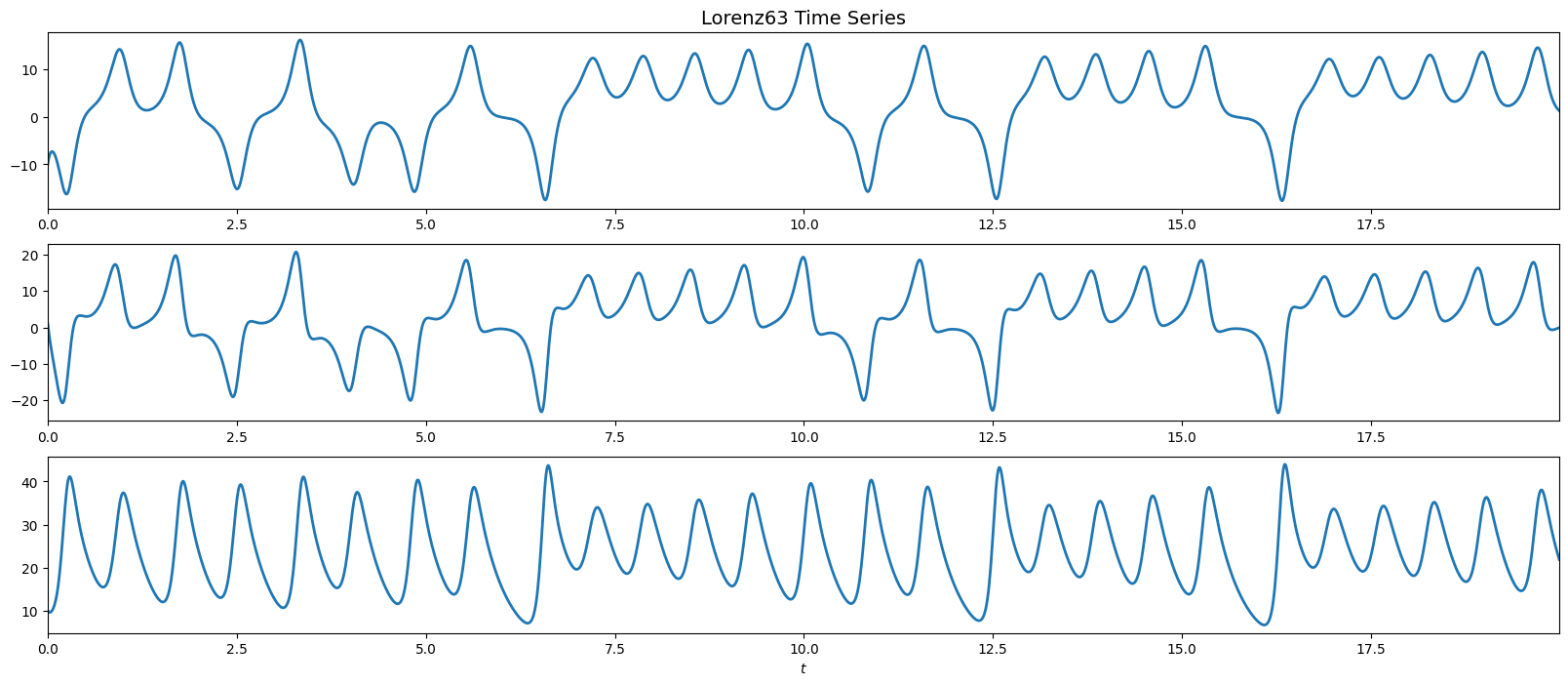}}

\textbf{Rössler} --- 3D chaotic system with simpler attractor geometry
than Lorenz.

\begin{lstlisting}[style=pythoncode]
u,t = orc.data.rossler(tN = 100, dt = 0.01)
vis.plot_time_series(u,t, title="Rossler Time Series")
\end{lstlisting}

\pandocbounded{\includegraphics[keepaspectratio]{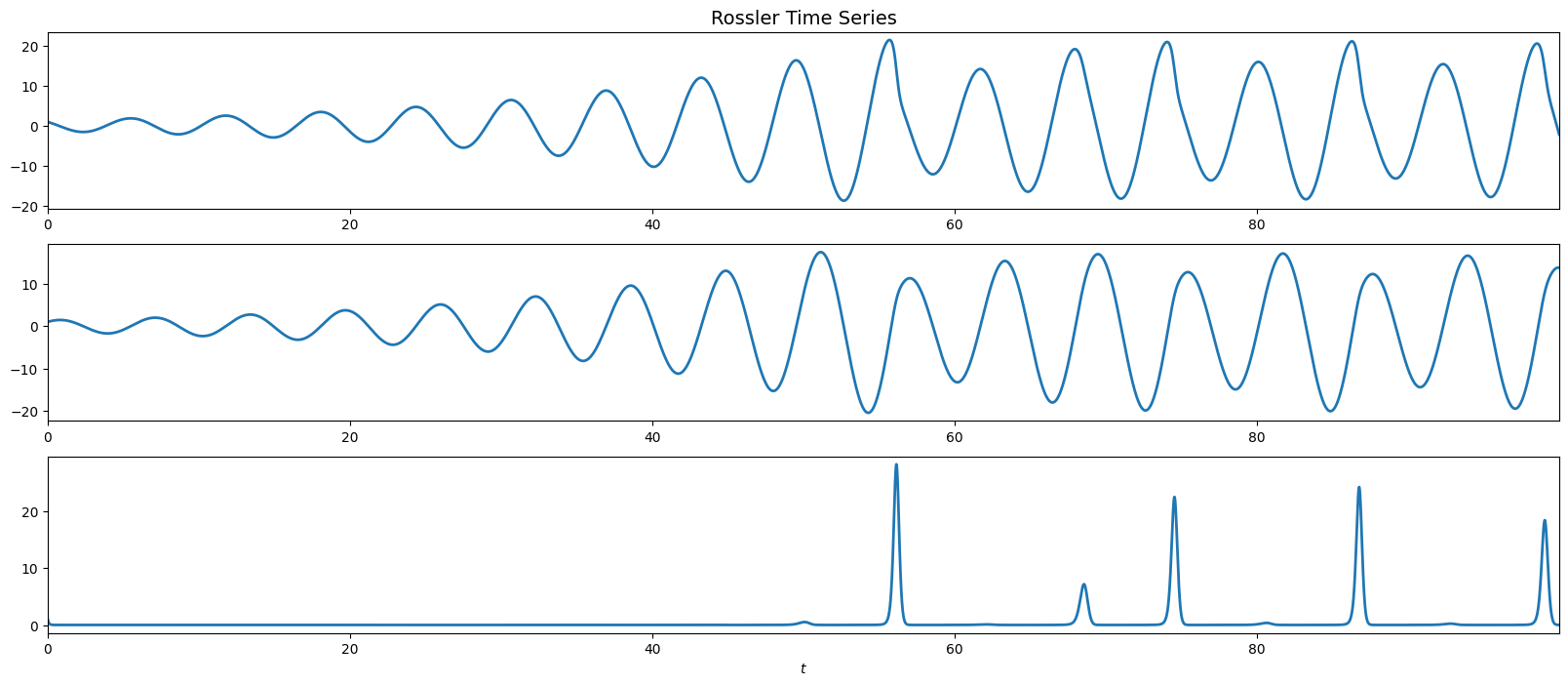}}

\textbf{Sakaraya} --- 3D chaotic system from ecological modeling
(predator-prey dynamics).

\begin{lstlisting}[style=pythoncode]
u,t = orc.data.sakaraya(tN = 20, dt = 0.01)
vis.plot_time_series(u,t, title="Sakaraya Time Series")
\end{lstlisting}

\pandocbounded{\includegraphics[keepaspectratio]{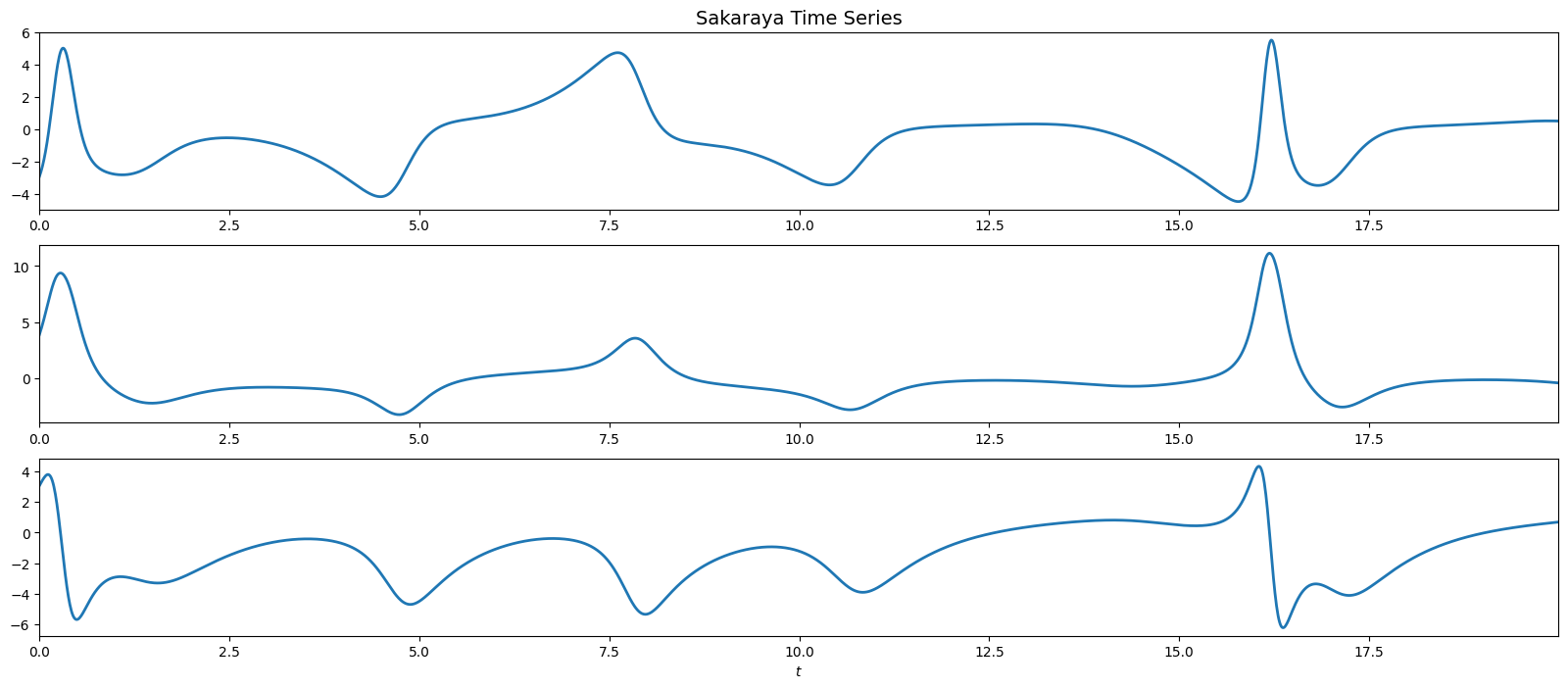}}

\textbf{Colpitts} --- 3D chaotic oscillator originating from electronic
circuit theory.

\begin{lstlisting}[style=pythoncode]
u,t = orc.data.colpitts(tN = 100, dt = 0.01)
vis.plot_time_series(u,t, title="Colpitts Time Series")
\end{lstlisting}

\pandocbounded{\includegraphics[keepaspectratio]{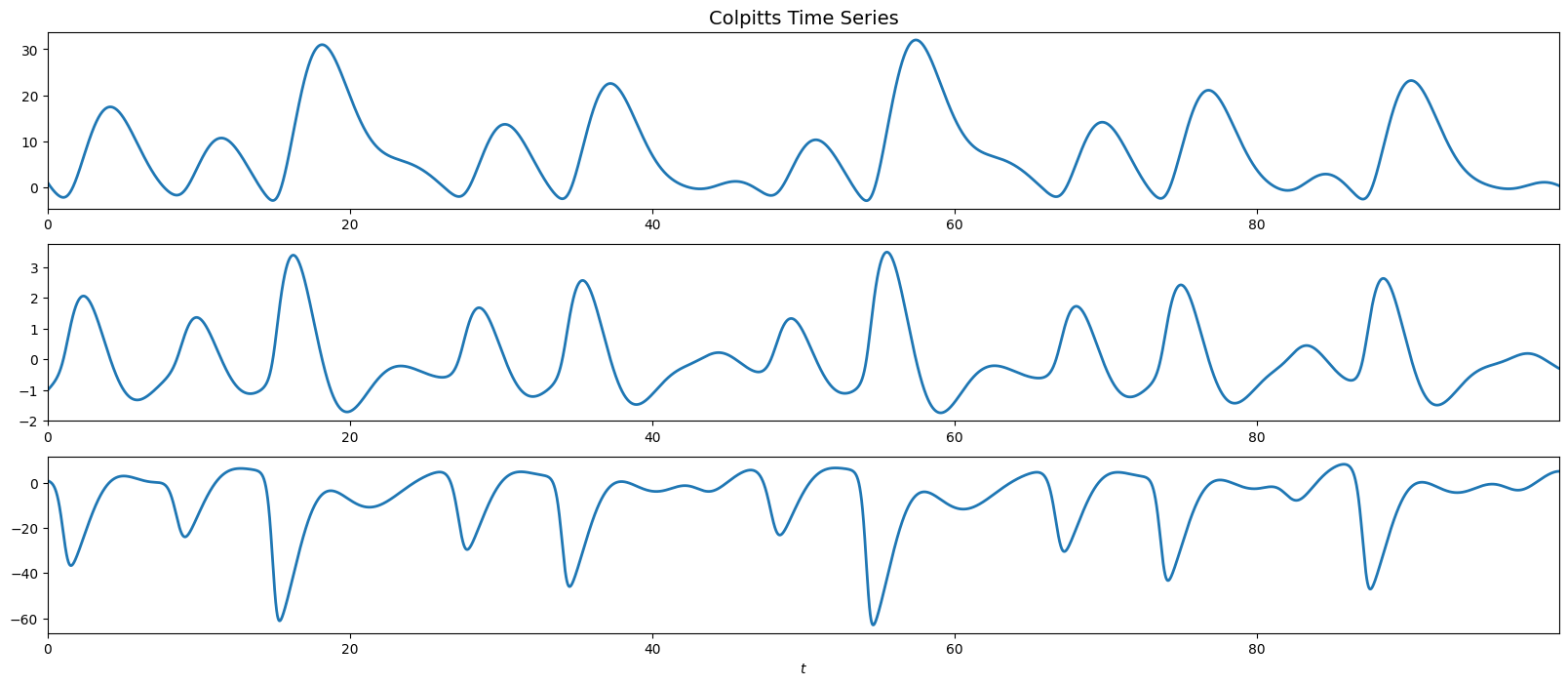}}

\textbf{Hyper Lorenz63} --- 4D hyperchaotic extension of the Lorenz
system (two positive Lyapunov exponents).

\begin{lstlisting}[style=pythoncode]
u,t = orc.data.hyper_lorenz63(tN = 20, dt = 0.01)
vis.plot_time_series(u,t, title="Hyper Lorenz63 Time Series")
\end{lstlisting}

\pandocbounded{\includegraphics[keepaspectratio]{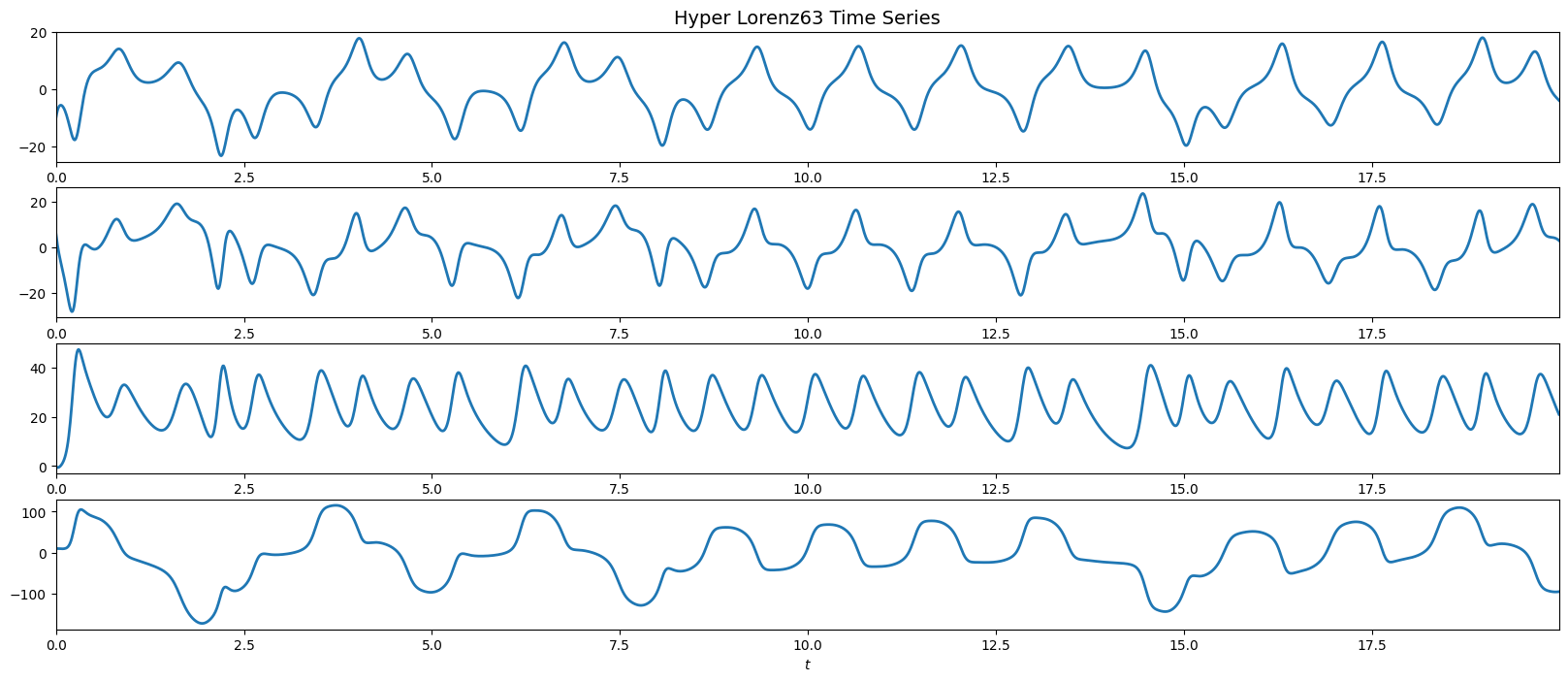}}

\textbf{Hyper Xu} --- 4D hyperchaotic system with complex attractor
structure.

\begin{lstlisting}[style=pythoncode]
u,t = orc.data.hyper_xu(tN = 20, dt = 0.01)
vis.plot_time_series(u,t, title="Hyper Xu Time Series")
\end{lstlisting}

\pandocbounded{\includegraphics[keepaspectratio]{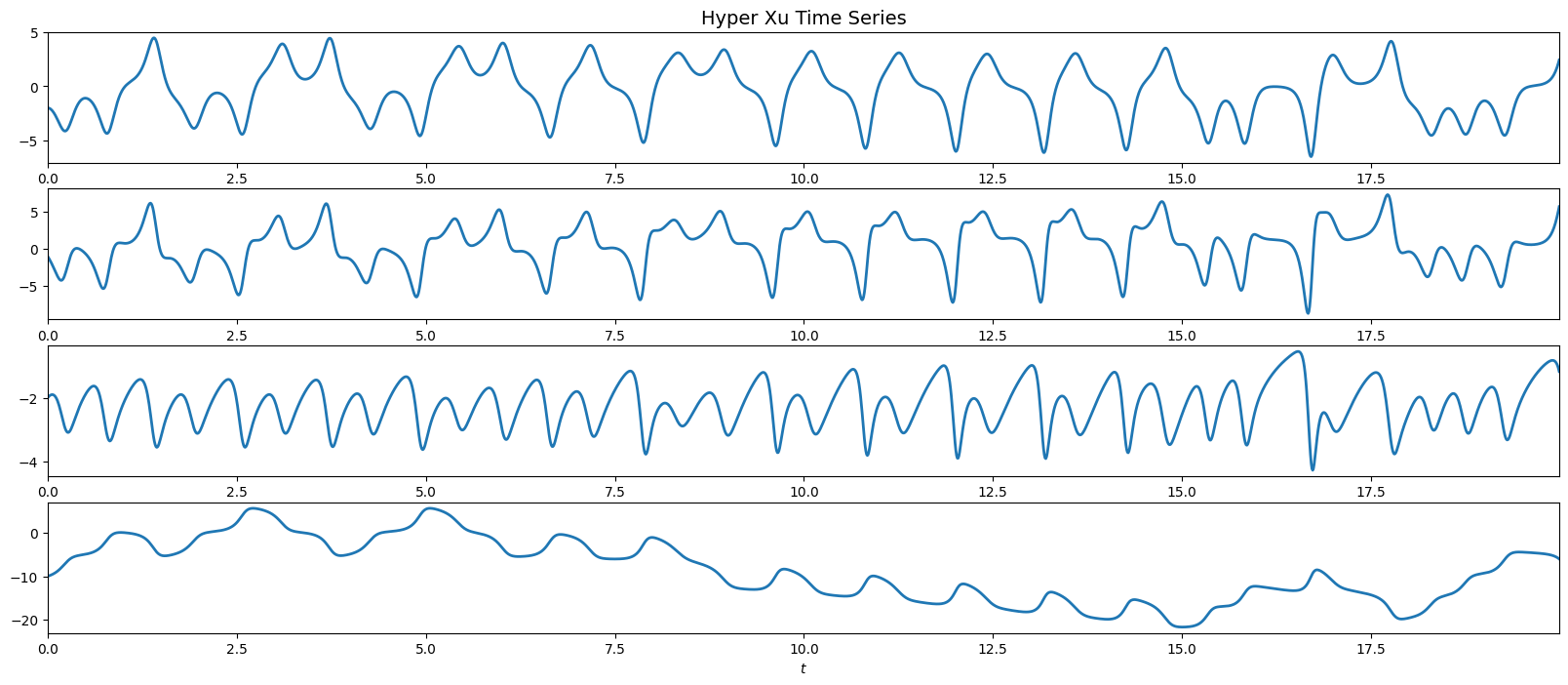}}

\textbf{Double Pendulum} --- 4D Hamiltonian system exhibiting chaos.
Supports optional damping.

\begin{lstlisting}[style=pythoncode]
u,t = orc.data.double_pendulum(tN = 40, dt = 0.01, damping=0.0)
vis.plot_time_series(u,t, title="Double Pendulum Time Series")
\end{lstlisting}

\pandocbounded{\includegraphics[keepaspectratio]{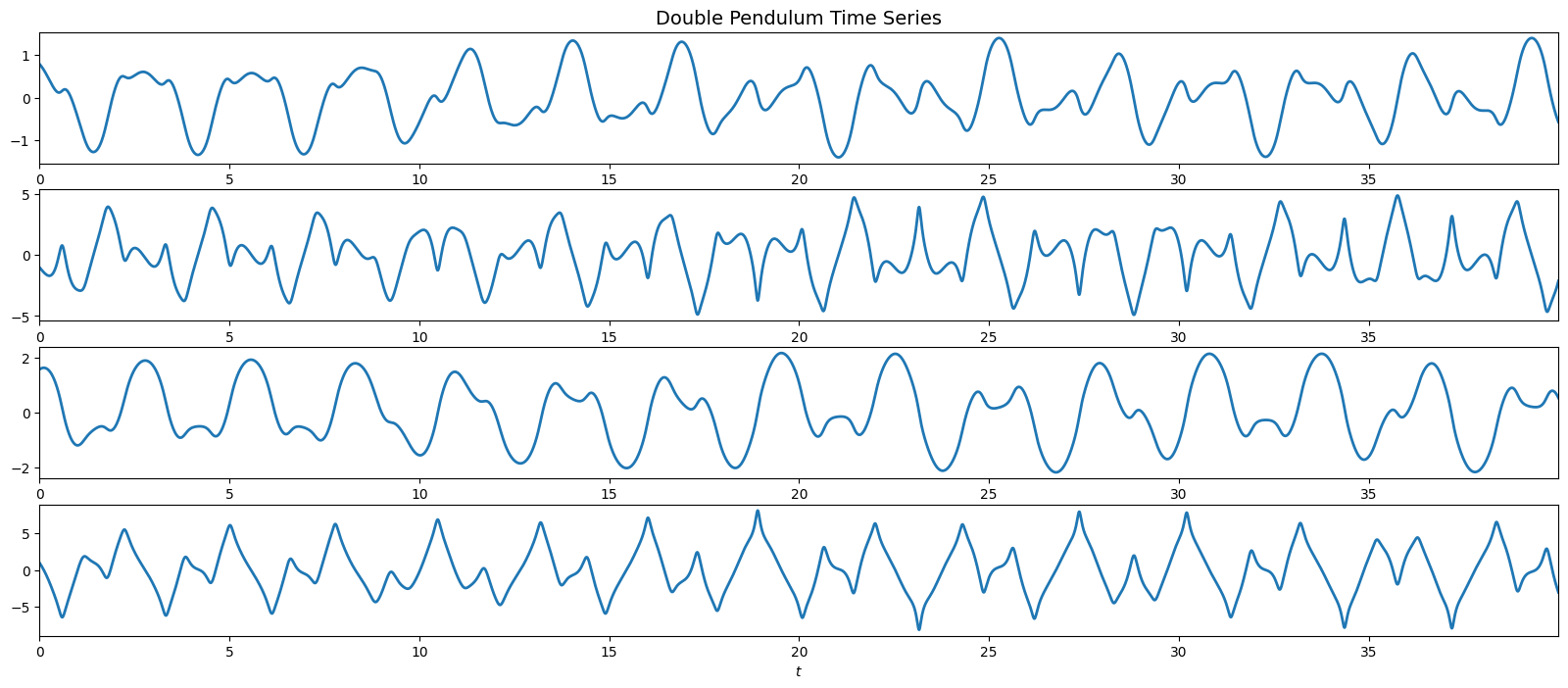}}

\subsection{Spatiotemporal Systems}\label{spatiotemporal-systems}

\textbf{Lorenz96} --- \(N\)-dimensional spatiotemporal chaos with
adjustable dimension. Commonly used for testing high-dimensional
forecasting methods.

\begin{lstlisting}[style=pythoncode]
u,t = orc.data.lorenz96(tN = 40, dt = 0.05, N=200)
vis.imshow_1D_spatiotemp(
    u,
    t[-1],
    title="Lorenz96 Time Series",
    interpolation='bicubic'
    )
\end{lstlisting}

\pandocbounded{\includegraphics[keepaspectratio]{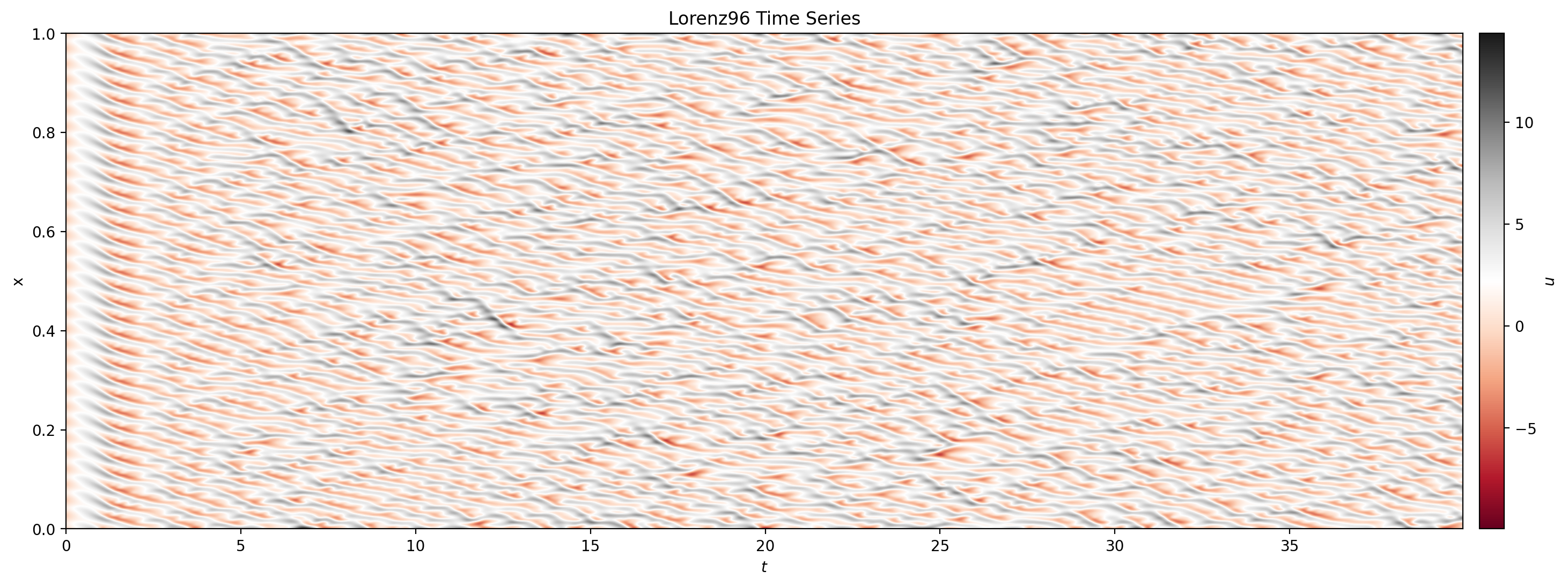}}

\textbf{Kuramoto-Sivashinsky (1D)} --- Spatiotemporally chaotic PDE
integrated with a spectral method. A standard benchmark for
high-dimensional RC forecasting (see \texttt{ks.ipynb}).

\begin{lstlisting}[style=pythoncode]
u,t = orc.data.KS_1D(tN=1000)
vis.imshow_1D_spatiotemp(
    u,
    t[-1],
    title="Kuramoto-Sivashinsky 1D Time Series",
    interpolation='bicubic'
)
\end{lstlisting}

\pandocbounded{\includegraphics[keepaspectratio]{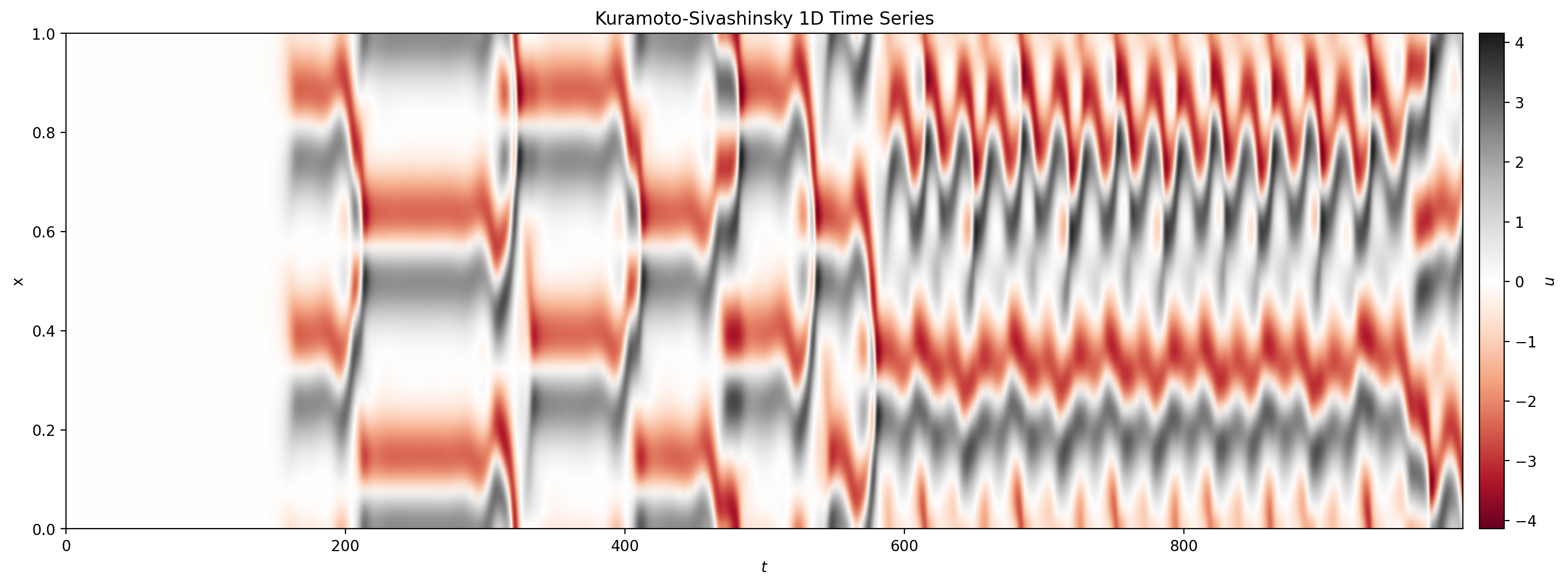}}

\end{document}